\pdfoutput=1

\documentclass[11pt]{article}

\usepackage[preprint]{acl}

\usepackage{microtype}

\usepackage{inconsolata}
\usepackage{booktabs}
\usepackage{titlesec}
\usepackage{multirow}
\usepackage{booktabs} 
\usepackage{tablefootnote} 
\usepackage{amsfonts}
\usepackage{float}
\usepackage{amsmath}
\usepackage{graphicx}
\usepackage{subcaption}
\usepackage{xcolor}

\usepackage{array}

\usepackage{times}
\usepackage{latexsym}
\usepackage{multirow}
\usepackage{booktabs}
\usepackage{graphicx}
\usepackage{amssymb}
\usepackage{enumitem}
\usepackage{amsmath}
\usepackage{float}
\usepackage{subcaption}
\usepackage[T1]{fontenc}

\usepackage[utf8]{inputenc}

\usepackage{microtype}

\usepackage{inconsolata}

\definecolor{agreen}{rgb}{0.55, 0.71, 0.0}
 \definecolor{lightcoral}{rgb}{0.94, 0.5, 0.5}
 
\newcommand{\vi}[1]{\textcolor{black}{#1}}
\newcommand{\vic}[1]{\textcolor{black}{#1}}

\newcommand{\xt}[1]{\textcolor{black}{#1}}

\newcommand{\inlineSubsection}[1]{
  \par\noindent\textbf{#1}\quad
}

\title{ \vi{An Empirical Study on Parameter-Efficient Fine-Tuning for \xt{MultiModal} Large Language Models}}

\author{
    Xiongtao Zhou$^1$\footnotemark[1] \quad 
    Jie He$^2$\thanks{\ \ Equal Contribution.} \quad 
    Yuhua Ke$^2$ \quad 
    Guangyao Zhu$^1$ \\
    \textbf{Víctor Gutiérrez-Basulto}$^3$ \, and \;
    \textbf{Jeff Z. Pan}$^2$\thanks{ \, Corresponding  author} \\
    $^1$ Waseda University, Japan \\
    $^2$ School of Informatics, University of Edinburgh, UK \\
    $^3$ School of Computer Science and Informatics, Cardiff University, UK \\
    \normalsize{\texttt{alenai.tao@ruri.waseda.jp, j.he@ed.ac.uk}} \\
    \normalsize{\texttt{s2484588@ed.ac.uk, zhuzgy@akane.waseda.jp}} \\
    \normalsize{\texttt{gutierrezbasultov@cardiff.ac.uk, j.z.pan@ed.ac.uk}} \\
}

\begin{document}
\maketitle
\begin{abstract}
\vi{\vic{Multimodal large language models (MLLMs) fine-tuned with multimodal instruction  datasets have demonstrated remarkable capabilities in multimodal tasks. However, fine-tuning all parameters of MLLMs has become challenging  as they  usually contain billions of parameters. To address this issue, we study \emph{parameter-efficient fine-tuning} (PEFT) methods for MLLMs.  We aim to identify effective methods for enhancing the performance of MLLMs in scenarios where only a limited number of parameters are trained. This paper conducts empirical studies using four popular PEFT methods to fine-tune the  LLM component of open-source MLLMs. We present a comprehensive analysis that encompasses various aspects, including the impact of PEFT methods on various models, parameters and location of the PEFT module, size of fine-tuning data, model stability based on PEFT methods, MLLM's generalization, and hallucination. We evaluated four PEFT methods on seven datasets from two different categories: unseen and seen datasets. Across all experiments, we show that the adapter is the best-performing PEFT method. At the same time, fine-tuning the connector layers  leads to improved performance in most MLLMs. Code and data are available at \href{https://github.com/alenai97/PEFT-MLLM.git}{https://github.com/alenai97/PEFT-MLLM.git} }}


\end{abstract}

\section{Introduction}
\vi{\vic{In recent years, the landscape of multimodal learning has been transformed by the emergence of multimodal large language models (MLLMs), such as LLaVA \cite{Liu2023VisualIT}, 
 MiniGPT4 \cite{zhu2024minigpt}, and GPT4-Vision \cite{openai2023gpt4}. MLLMs have showcased impressive competency across a spectrum of multimodal benchmarks \cite{fu2023mme,liu2023mmbench,Li2023SEEDBenchBM} 
 thanks to \xt{t}he integrated architecture of pre-trained visual encoders, connector layers, and LLMs. This architecture is usually fine-tuned through multimodal instruction-following data \cite{xu-etal-2023-multiinstruct}.}}
\vic{During fine-tuning, most existing MLLMs \cite{cha2023honeybee,su2023pandagpt,lin2023sphinx} typically freeze the visual encoder, focusing solely on connector layers and the LLM component. Since LLMs (e.g.\  LLaMA \cite{touvron2023llama} and Vicuna-v1.5 \cite{vicuna2023}) often contain hundreds of billions of parameters,   \emph{full fine-tuning} (FFT) \cite{wang-etal-2022-adamix} is unfeasible. Consequently,  the \emph{parameter-efficient fine-tuning} (PEFT) \cite{pmlr-v97-houlsby19a,hu2021lora} approach (which leverages lightweight trainable parameters  and keeps  the majority of parameters  frozen) has been widely employed in NLP for  fine-tuning LLMs with  instruction or task-specific datasets \cite{li2023LLaVAmed,you2023ferret}, as they allow for significant resource savings while achieving comparable performance or even surpassing FFT \cite{peft}.}


\vic{
In contrast to standard LLMs, MLLMs introduce additional modules: visual encoder and connector layers. During the fine-tuning process, unimodal LLMs only receive text features while MLLMs get multimodal inputs, {such that connector layers are also fine-tuned, not just fine-tuning the LLM.} Therefore, it is crucial to reassess the performance of fine-tuning MLLMs using various PEFT methods, exploring the impact of connector fine-tuning on the model's performance in downstream tasks, and examining PEFT's effects on model stability, generalization, and hallucination.}
%
\vic{In this paper we address these issues by conducting comprehensive studies on three representative MLLMs containing connector layers}: LLaVA-1.5 (7B, 13B)  \cite{liu2023improved}, ShareGPTv4 (7B)  \cite{chen2023sharegpt4v}, and Qwen-VL-Chat (7B)  \cite{bai2023qwenvl}.
\vic{Our study looks at various issues related to PEFT methods. Specifically, we design our study to address the following questions: \textbf{(1)} \vic{Is it necessary to fine-tune the connector} when fine-tuning  MLLMS via various PEFT methods  on unseen and seen datasets? \textbf{(2)} How does the position of the PEFT module in the LLM affect the MLLM's performance? \textbf{(3)} Faced with different training data scales, what differences exist in the performance of different PEFT methods? \textbf{(4)} How do different PEFT approaches impact the stability of the model? Is there any relationship between trainable parameters and learning rate with stability?}

\begin{figure*}
    \centering
    \includegraphics[width=1\linewidth]{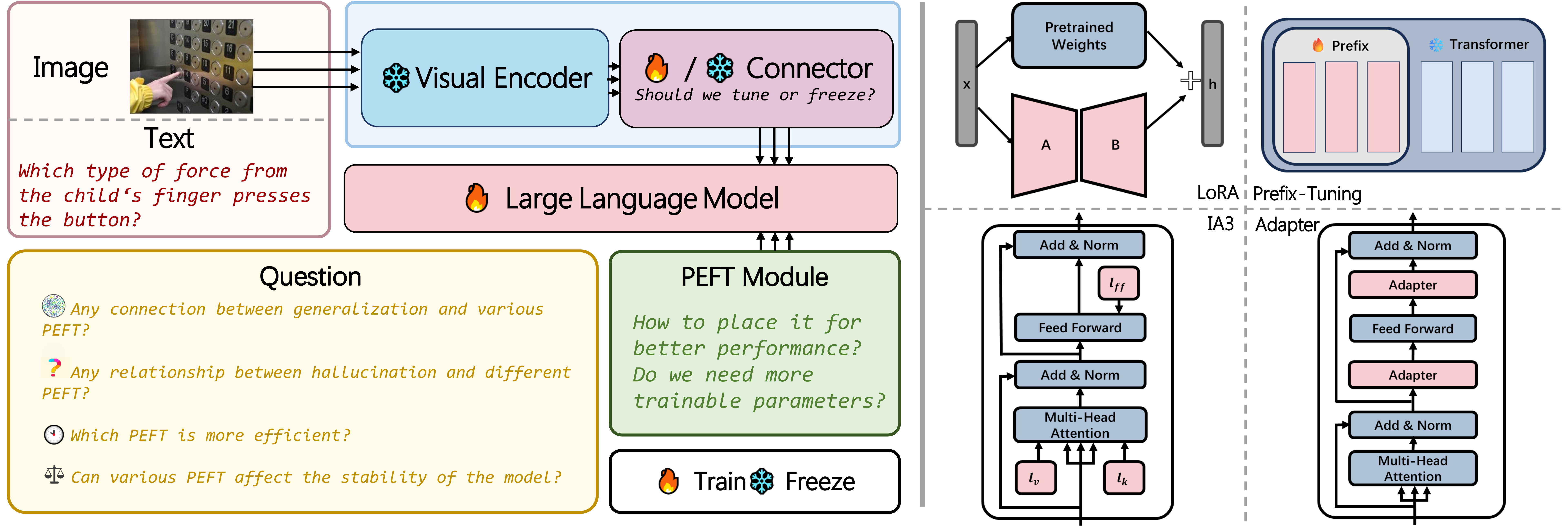}
    \caption{Left): Architecture of a Multimodal Large Language Model. Starting from 7 questions, we comprehensively explored the impact of PEFT methods and the connector on MLLMs, all of which are illustrated on the Left. Right): A detailed illustration of the PEFT module structure for the four PEFT methods.}
    \label{fig:main_fig}
\end{figure*}

Our \textbf{key findings} can be summarized as follows:

\begin{enumerate}
\setlength{\itemsep}{0pt}
\setlength{\parsep}{0pt}
\setlength{\parskip}{0pt}
    \item \vic{Fine-tuning the connector layers usually leads to   performance improvement within MLLMs.} 
    \item \vic{More trainable parameters results in better performance on unseen datasets, while fewer trainable parameters maintains the model's performance on seen datasets.}
    \item \vic{Generally,   fine-tuning using large scale datasets leads to better performance.  However,  when resources are limited, 
    } one should consider medium-size datasets instead.
    \item \vic{Adapters show the best overall performance in model generalization, stability, and hallucination.}
\end{enumerate}
\vic{Our contributions can be summarized as follows:
{(1) We have assembled a standardized evaluation suite  that includes seven benchmarks from the vision-and-language research community.} This suite encompasses five tasks in visual question answering, one in visual reasoning, and one in image caption, along with four PEFT methods. \textbf{(2)} We utilized these resources to conduct {in-depth experiments investigating four crucial design dimensions} (cf.\ Fig.\ \ref{fig:main_fig}, left): 1) data scaling, 2) stability of the training process, 3) overfitting and generalization, and 4) hallucination. {(3)} Our empirical findings show that {Adapter outperforms other PEFT methods in all aspects}, followed in second place by LoRA. Furthermore, we show  that {fine-tuning the connector layers frequently enhances performance within MLLMs.}}
\section{Related Work}
\inlineSubsection{\vi{Multimodal Large Language Models.}}
\vic{Flamingo~\cite{alayrac2022flamingo} proposes a GATED XATTN-DENSE layer to align visual and textual features,  connecting the visual module and language model.  
LLaMA-adapter~\cite{zhang2024llamaadapter} applies a projection layer to connect a visual encoder and LLaMA. It proposes adding an adapter module on  LLaMA, keeping only the adapter parameters updated during training. In contrast, LLaVA-1.5 
 \cite{liu2023improved} employs two layers of MLP to connect the visual encoder and LLM. During fine-tuning, it only updates the parameters of  the MLP and LLM. Subsequent works  mostly build upon this approach, employing the connector layers to link a visual encoder and LLMs, and then fine-tune the model using multimodal instruction-following data \cite{li2023otter,hu2023bliva,wang2023cogvlm}.} Recently, there are also many work on multimodal large language models~\cite{CZFG+2024} from the perspective of knowledge computing~\cite{PRKSC2023}.

\inlineSubsection{Parameter-Efficient Fine-Tuning.} 
\vic{Parameter-efficient fine-tuning emerges as an approach capable of achieving performance comparable to full fine-tuning while keeping the majority of parameters frozen. Prompt-based methods~ \cite{lester-etal-2021-power} incorporate soft prompts into the input prefixes, only updating these soft prompts.
Another widely used family of methods is based on adapters~\cite{pfeiffer-etal-2020-mad,he2022towards,pmlr-v203-he23a}, which insert adapter modules at specific positions within transformer layers and update only the parameters of these inserted modules during training. Also, in MLLMs, low-rank decomposition methods are commonly employed~\cite{hu2021lora,edalati2022krona}. These methods involve training only the parameters in low-rank matrices, significantly reducing the number of trainable parameters. 
}

\section{PEFT Methods}
\vic{Figure~\ref{fig:main_fig} illustrates the architecture of MLLMs and the location of various PEFT modules. In our experiments, all considered MLLMs consist of three components: a visual encoder, connector layers, and a LLM. The structure of the connector layers may vary depending on the specific MLLM.  PEFT methods can be classified into three categories, from which we select four methods: (1) \emph{reparametrization-based tuning}: LoRA, IA3  (2) \emph{adapter-based tuning}: Adapter. (3) \emph{prompt-based tuning}: Prefix-Tuning.}



\inlineSubsection{LoRA.} \vic{We integrate the low-rank strategy proposed by \citet{hu2021lora} to adjust the network weights, facilitating the model's handling of complex tasks with an efficient parameter footprint. The original pre-trained weight matrix $W_0\in{\mathbb{R}^{d\times{k}}}$ is updated through low-rank decomposition using Equation \ref{equ: LoRA}, where $B\in{\mathbb{R}^{d\times{r}}}$ and $A\in{\mathbb{R}^{r\times{k}}}$. }
\begin{equation}
    W_0 + \Delta W = W_0 + BA
    \label{equ: LoRA}
\end{equation}
\vic{This method ensures our model's adaptability is improved without a significant increase in the parameter space.}

\inlineSubsection{IA3.} 
\vic{Following \citet{liu2022fewshot}, we integrate three vectors $v_{k}\in{\mathbb{R}^{d_k}}$, $v_{v}\in{\mathbb{R}^{d_v}}$, and $v_{f\!f}\in{\mathbb{R}^{d_{f\!f}}}$ into an attention mechanisms as:}
\begin{equation}
    {\rm{softmax}}(\frac{Q(v_k\odot{K^T})}{\sqrt{d_k}})(v_v\odot V)
\end{equation}
\xt{where $\odot$ represents the element-wise multiplication, }\vic{and ($v_{f\!f} \odot \gamma(W_1x))W_2$ in the position-wise FFN layers, leveraging $\gamma$ as the activation function. These formulas guide the model's  attention to be fine-tuned to prioritize relevant features, optimizing performance without significantly increasing the model's complexity or number of parameters.}

\inlineSubsection{Adapter.} \vic{We adopt the structure proposed by \citet{pmlr-v97-houlsby19a}, which adds adapter modules to the fully-connected networks after attention and the FFN layer within the transformer layers. This can be captured as follows:}
\begin{equation}
     h_i + f(W_{down}(h_i))W_{up} \rightarrow h_i
\end{equation}
\vic{where $h_i$ is the output of the previous layer, which is initially down-projected by $W_{\text{down}} \in \mathbb{R}^{d \times r}$ to a lower dimension $r$, and then up-projected back by  $W_{\text{up}} \in \mathbb{R}^{r \times d}$  to the original dimension $d$, $f$ is a non-linear layer.}

\inlineSubsection{Prefix-Tuning.}\vic{We follow the approach proposed by \citet{li-liang-2021-prefix} to employ prefix learning by appending task-specific vector ``prefixes'' to the input sequence fed into the pre-trained model. We initialize a trainable matrix $P_{\theta}$ with dimensions $|P_{idx}|\times {\rm dim}(y_i)$, where $P_{idx}$ specifies the prefix length. This yields the following conditional formulation for each element $y_i$ of the output sequence:}
\begin{equation}
y_i = 
\begin{cases} 
P_{\theta}[i, :] & \text{if } i \in P_{{idx}}, \\
LM_{\phi}(z_i, y_{<i}) & \text{otherwise}.
\end{cases}
\end{equation}
\xt{If $i \in P_{{idx}}$, a bidirectional encoder computes the $y_i$. For $i \notin P_{{idx}}$, $y_i$ is computed by an autoregressive neural LM as a function of $y_i$ and the past activations in its left context.}

\section{Experiment Setup}

\subsection{Datasets}

\vic{In the current era of large-scale models, dataset contamination is a significant concern as it is challenging to ensure that the data will be used for the next training process constitutes unseen data for large language models. Therefore,  we categorize the datasets into two types: \textbf{Unseen datasets}, comprising datasets that have not been involved in the training of any of the considered models,  including (1) the ScienceQA dataset~\cite{lu2022learn}; (2) the Vizwiz dataset~\cite{Gurari_2018_CVPR}; (3) the IconQA dataset~\cite{lu2021iconqa}; and (4) the Flickr30k dataset~\cite{young-etal-2014-image}. \textbf{Seen datasets}, consisting of datasets used in the training of all considered models, including (1) the OKVQA dataset~\cite{okvqa}; (2) the OCRVQA dataset~\cite{mishraICDAR19}; and (3) the VQAv2 dataset~\cite{balanced_vqa_v2}. Details about datasets
can be found in App.~\ref{appendix_A}.}

\subsection{Implementations}

\inlineSubsection{Models.}  \vic{We selected LLaVA-1.5 (7B, 13B), ShareGPTv4 (7B), and Qwen-VL-Chat (7B) as the base models for our experiments.}


\inlineSubsection{Hyperparameters.}
\vic{ We conduct fine-tuning on the  training set of each dataset separately and then test on their respective test or validation sets. All experiments were conducted with a global batch size of 128. We set the random seed of the experiment to 42. Additionally, each PEFT method was trained for three epochs on the fine-tuning dataset. For LoRA, we set its learning rate to 2e-4, the adapter's to 5e-5, IA3's to 2e-4, and Prefix-Tuning's to 1e-5. More information about model and hyperparameter settings is available in  App.~\ref{appendix_B} }

\section{\vi{Experimental Results}}
\subsection{\vi{Main Results}}

\begin{table*}[t]
\scriptsize
  \centering
  \begin{tabular}{clcccccccccc}
  \toprule
  Model & Method & SQA (img) & VizWiz & IconQA-txt & IconQA-blank & Flickr30k & OKVQA & OCRVQA & VQAv2 & Avg \\
  \midrule
  \multirow{8}{*}{LLaVA-1.5-7B} 
      & Adapter         & 78.7  & 66.7  & 83.5  & 77.7  & 91.1  & 59.4  & 65.5  & 74.0  & 74.6 \\
      & -w/ connector   & 84.4  & 67.6  & 88.7  & 80.9  & 89.8  & 59.8  & 65.2  & 73.8  & 76.3 \\
      & LoRA            & 85.2  & 64.7  & 89.9  & 85.5  & 85.6  & 56.3  & 68.2  & 73.2  & 76.1 \\
      & -w/ connector   & 86.2  & 66.5  & 90.6  & 88.8  & 85.2  & 56.5  & 66.7  & 73.1  & 76.7  \\
      & IA3             & 69.1  & 56.2  & 55.2  & 41.9  & 77.2  & 59.8  & 66.9  & 78.1  & 63.1  \\
      & -w/ connector   & 82.7  & 61.9  & 89.2  & 82.2  & 91.9  & 60.5  & 67.1  & 75.2  & 76.3 \\
      & Prefix          & 68.2  & 59.0  & 73.0  & 46.8  & 91.5  & 61.1  & 68.6  & 76.9  & 68.1 \\
      & -w/ connector   & 69.7  & 60.8  & 76.7  & 50.9  & 91.9  & 61.3  & 68.5  & 77.0  & 69.6 \\

  \midrule
    \multirow{8}{*}{LLaVA-1.5-13B}
      & Adapter         & 82.4  & 66.6  & 88.9  & 84.2  & 94.0 & 59.4  & 67.4  & 74.7  & 77.2 \\
      & -w/ connector   & 83.7  & 66.8  & 90.6  & 85.8  & 93.1 & 59.6  & 67.2  & 74.5  & 77.7 \\
      & LoRA            & 86.3  & 66.3  & 90.9  & 90.3  & 87.9 & 59.1  & 70.8  & 74.4  & 78.3 \\
      & -w/ connector   & \textbf{87.8}  & 66.1  & 91.6  & 90.4  & 84.1 & 59.9  & 68.6  & 73.6  & 77.8 \\
      & IA3             & 72.3  & 58.8  & 58.9  & 47.5  & 70.9 & 62.6  & 70.5  & 78.4  & 65.0 \\
      & -w/ connector   & 84.5  & 67.3  & 90.3  & 84.8  & 91.3 & 63.8  & 69.0    & 76.7  & \textbf{78.5} \\
      & Prefix          & 70.4  & 68.7  & 65.2  & 41.5  & 88.2 & 64.4  & 66.8  & 77.9  & 67.9 \\
      & -w/ connector   & 71.7  & 69.1  & 65.7  & 46.8  & 89.1 & \textbf{64.7}  & 67.4  & 78.6  & 69.1 \\

  \midrule
  \multirow{8}{*}{ShareGPT4V} 
      & Adapter         & 81.1  & 67.0  & 89.7  & 82.8  & \textbf{95.6}  & 59.8  & 67.9  & 76.7  & 77.6 \\
      & -w/ connector   & 82.2  & 64.1  & 91.8  & 86.0  & 93.4  & 59.5  & 67.5  & 76.2  & 77.6 \\
      & LoRA            & 86.7  & 65.6  & 91.8  & 90.4  & 85.0  & 57.9  & 69.8  & 75.9  & 77.9 \\
      & -w/ connector   & 86.7  & 67.3  & 91.9  & \textbf{90.6}  & 84.9  & 57.6  & 69.0  & 75.3  & 77.9 \\
      & IA3             & 69.0  & 61.1  & 58.7  & 47.7  & 57.5  & 60.7  & 69.1  & \textbf{79.8}  & 63.0 \\
      & -w/ connector   & 82.0  & 60.9  & 90.9  & 84.1  & 93.8  & 61.4  & 68.7  & 77.3  & 77.4 \\
      & Prefix          & 67.9  & 63.6  & 73.8  & 45.2  & 91.6  & 62.4  & 68.9  & 78.7  & 69.0 \\
      & -w/ connector   & 68.4  & 65.2  & 81.3  & 53.2  & 92.4  & 62.3  & 67.7  & 78.8  & 71.2 \\

  \midrule
  \multirow{8}{*}{Qwen-VL-Chat} 
      & Adapter         & 79.6  & 67.8  & \textbf{92.4} & 90.5  & 86.4  & 54.9  & 71.1  & 75.8  & 77.3 \\
      & -w/ connector   & 81.2  & 69.3  & 90.8  & 87.5  & 82.7  & 51.1  & 69.3  & 70.7  & 75.3 \\
      & LoRA            & 86.8  & 68.5  & 91.5  & 85.5  & 82.6  & 53.8  & \textbf{71.4}  & 75.7  & 77.0 \\
      & -w/ connector   & 84.0  & 68.8  & 71.9  & 90.3  & 83.5  & 43.5  & 67.0  & 63.3  & 71.5 \\
      & IA3             & 70.0  & 66.9  & 71.0  & 41.8  & 73.6  & 50.7  & 68.3  & 77.8  & 65.0 \\
      & -w/ connector   & 67.3  & 69.8  & 57.3  & 28.7  & 65.1  & 50.5  & 62.1  & 77.5  & 59.8 \\
      & Prefix          & 52.2  & \textbf{70.6}  & 52.4  & 33.2  & 52.2  & 50.1  & 61.3  & 70.6  & 55.3 \\
      & -w/ connector   & 51.9  & 70.4  & 52.5  & 31.8  & 52.9  & 49.8  & 61.5  & 77.4  & 56.0 \\

  \bottomrule
  \end{tabular}%
  \caption{Main experimental results of various MLLMs with four PEFT methods. w/ connector: Tuning the connector.}
  \label{tab:main}
\end{table*}

\begin{figure}[t]
    \centering
    \begin{subfigure}{\columnwidth}
           \centering
 \includegraphics[width=0.75\columnwidth]{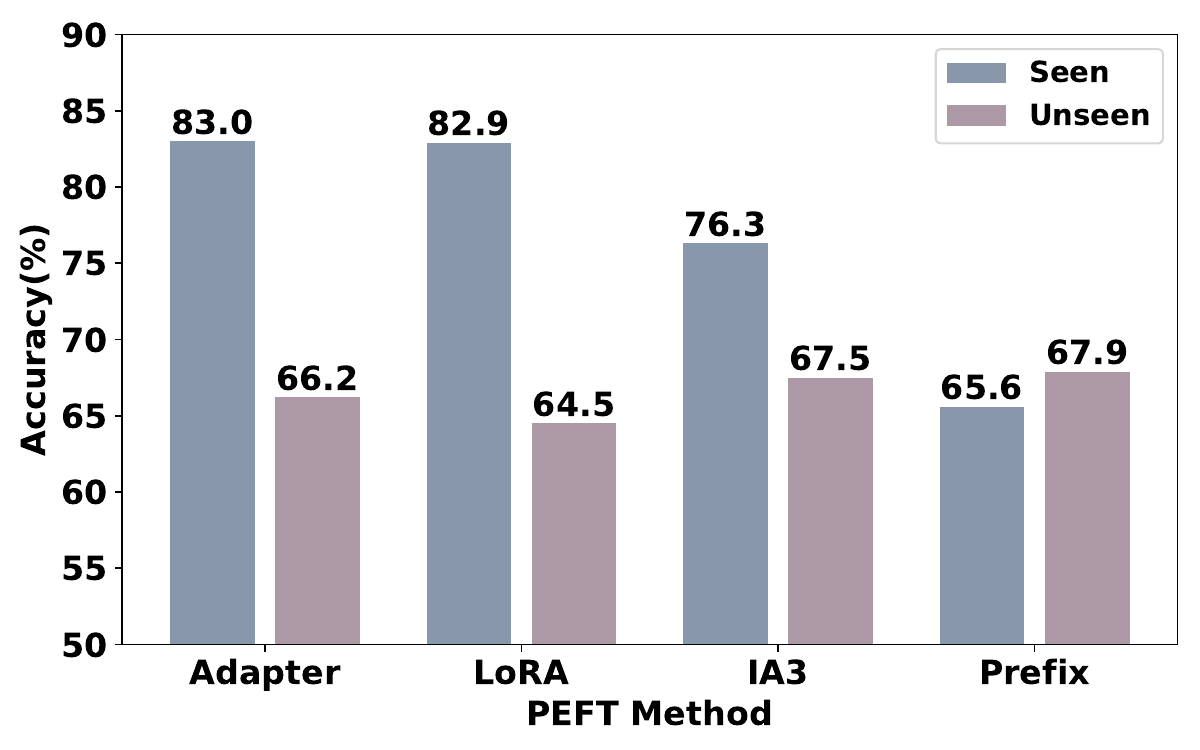}
        \caption{w/ connector}
        \label{fig:connector}
    \end{subfigure}
    
    \hfill
    \begin{subfigure}{\columnwidth}
            \centering
\includegraphics[width=0.75\columnwidth]{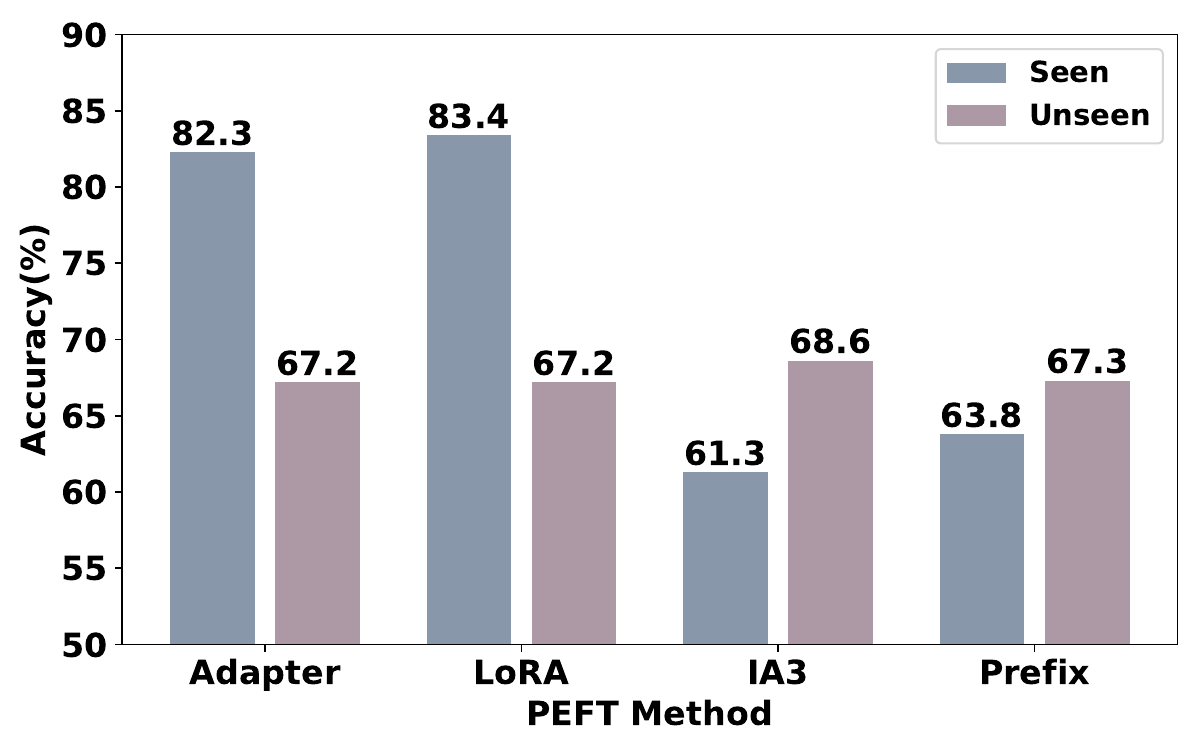}
        \caption{w/o connector}
        \label{fig:no-connector}
    \end{subfigure}
    \caption{The comparative performance of four PEFT methods on seen and unseen datasets, with and without the use of a connector.}
    \label{fig:main}
\end{figure}

\xt{\emph{Should we tune or freeze the connector when considering unseen and seen datasets?}} \vic{Given the increasing availability of pretraining data for MLLMs, encountering contaminated data (i.e. training data contains information that is meant to be present only in the test set) is increasingly common. \vic{Additionally, most current MLLMs tune the connector layers during fine-tuning, yet the role of the connector remains unclear.} Our main experiment thus focuses on investigating the performance of  PEFT methods on both unseen and seen datasets. In our experiments, following existing work \cite{Liu2023VisualIT,bai2023qwenvl,chen2023sharegpt4v},  we freeze the visual encoder and apply the PEFT module to fine-tune the LLM. For the connector layers, we experimented with both FFT and freezing. \vic{We investigate the model's performance on certain tasks during the task-specific fine-tuning when unfreezing the visual encoder, see App.~\ref{app:add} for details about unfreezing the visual encoder.} We investigated the performance of LLaVA-1.5 (7B, 13B), ShareGPT4v, and Qwen-VL-Chat on \vic{the datasets mentioned in Section 4.1.} }

\vic{The obtained results are presented in Table~\ref{tab:main}. One can observe that LLaVA-1.5-13B with IA3 and fine-tuned connector layers achieved the best results across all unseen and seen datasets. \vic{Most IA3 models with fine-tuned connector achieved comparable performances to LoRA and Adapter on unseen datasets, while also maximizing the model's performance on seen datasets.}  \vic{Benefiting from the increased number of parameters in LLaVA-1.5-13B, we noticed that across various settings, the average performance on all datasets of LLaVA-1.5-13B surpasses that of LLaVA-1.5-7B.} \vic{The average performance of ShareGPTv4 generally surpasses (except for the IA3 method with the frozen connector) that of LLaVA-1.5-7B,}  because it has been fine-tuned on the higher-quality multimodal instruction-following dataset \cite{chen2023sharegpt4v}.  \vic{Under the setting of freezing the connector layers and fine-tuning the LLM with LoRA, Qwen-VL-Chat achieved the best performance. We found that choosing to tune the connector layers often leads to a significant deterioration in Qwen-VL-Chat's performance on seen datasets.}  Figure~\ref{fig:main} illustrates the performance of various PEFT methods under the settings of tuning or freezing the connector layers.
When fine-tuning the connector layers, LoRA, Prefix-Tuning, and IA3 all exhibit better performance \xt{than freezing the connector layers} on unseen datasets. In this case, IA3 gets a 15.0\% increase in the average result. On seen datasets, in most cases, the performance of freezing the connector layers and that of the remaining PEFT methods (except for a slight decrease in LoRA's performance) is similar. Note that whether the connector layers are fine-tuned or not, the performance of the Adapter remains relatively consistent on both seen and unseen datasets.}
\vi{Our \textbf{main findings} are the following:}
\begin{itemize}[itemsep=0.5ex, leftmargin=5mm]
\setlength{\itemsep}{0pt}
\setlength{\parsep}{0pt}
\setlength{\parskip}{0pt}
    \item \vic{LoRA and Adapter have the best performance on \xt{all of} the unseen datasets, while IA3 and Prefix-Tuning perform the best on the \xt{OKVQA and VQAv2}.} \vic{More trainable parameters allows the model to better adapt to unseen datasets, while fewer trainable parameters can maintain the model's performance on seen datasets.} 
    \item \vic{For the unseen datasets, tuning the connector layers often outperforms freezing the connector layers. For the seen datasets, freezing the connector layers yields the best performance. }
\end{itemize}



\subsection{Module Location}


\begin{table*}[htb]
\scriptsize
    \centering
    \begin{tabular}{llccccccccc}
    \toprule
          Method  & Location & SQA (img) & VizWiz & IconQA-txt & IconQA-blank & OKVQA & OCRVQA & VQAv2 & Avg \\
         \midrule
         \multicolumn{10}{c}{LLaVA-1.5$_{7B}$} \\
         \midrule
         \multirow{3}{*}{Adapter} 
         & Attn & 81.3  & 67.9  & 89.2  & 80.3  & 58.8  & 66.2  & 75.0 & 74.1 \\
         & MLP  & 84.4  & 67.6  & 88.7  & 80.9  & 59.8  & 65.2  & 73.8 & \textbf{74.3}\\
         & Both & 82.9  & 67.8  & 88.8  & 81.4  & 55.2  & 64.3  & 72.1 & 73.2\\
         \midrule
         \multirow{3}{*}{LoRA} 
         & Attn & 84.1  & \textbf{68.1}  & 90.5  & 83.8  & 58.4  & 67.0  & 73.5 & 75.1\\
         & MLP  & \textbf{85.6}  & 66.3  & \textbf{90.8}  & 88.0  & 56.5  & 66.5  & 73.0 & 75.2\\
         & Both & 86.2  & 66.5  & 90.6  & \textbf{88.8}  & 56.5  & 66.7  & 73.1 & \textbf{75.5}\\
         \midrule
         \multirow{3}{*}{IA3} 
         & Attn & 81.0  & 61.9  & 88.9  & 82.1  & 60.3  & 67.3  & 75.2 & 73.8\\
         & MLP  & 82.0  & 62.1  & 88.7  & 82.6  & \textbf{60.5}  & \textbf{67.4}  & \textbf{75.3} & \textbf{74.1}\\
         & Both & 82.7  & 61.9  & 89.2  & 82.2  & \textbf{60.5}  & 67.1  & 75.2 & \textbf{74.1}\\
    \bottomrule
    \end{tabular}%
    \caption{Average results of PEFT module location on LLaVA-1.5-7B. Attn: Placed on attention layer. MLP: Placed on MLP layer. Both: Placed both on attention layers and MLP layers.}
    \label{tab:main_locate}
\end{table*}

\vic{\emph{What is the best location for the PEFT module  for MLLMs?} Unlike LLMs, MLLMs include additional connector layers and visual encoders. Therefore, we can not straightforwardly transfer existing results for LLMs to MLLMs. With this in mind, we directly address this issue here.
To this end, we selected all VQA datasets for the location study. We choose LLaVA-1.5-7B as the base model set the random seed to 42, freeze the visual encoder, and fine-tune the connector layers and LLM. We used this setting in subsequent experiments. For LoRA and IA3, we integrate them into the model's multi-head attention layer, MLP layer, or both. For adapters, we placed them in the same locations. The result of Qwen-VL-Chat can be found in App.~\ref{app:C}.
Note that we do not consider Prefix-Tuning as the position is fixed.
Table~\ref{tab:main_locate} presents the results on LLaVA-1.5-7B, which suggest that despite the additional modules in MLLMs compared to LLMs, the results of \citet{hu-etal-2023-llm} for fine-tuning LLMs are also valid for MLLMs.}


\begin{itemize}[itemsep=0.5ex, leftmargin=5mm]
\setlength{\itemsep}{0pt}
\setlength{\parsep}{0pt}
\setlength{\parskip}{0pt}

    \item \vic{We observe that for LoRA and IA3, the \textbf{Both} setting achieved the best results. As for  Adapter, inserting it only into the \textbf{MLP} layer yielded the best performance.} 
\end{itemize}

\subsection{Data Scale}


\begin{table*}[htb]
\scriptsize
    \centering
    \begin{tabular}{llccccc|cccc}
    \toprule[1pt]
     & Method & SQA (img) & VizWiz & IconQA-txt & IconQA-blank & Flickr30k & OKVQA & OCRVQA & VQAv2 & Avg \\
    \midrule
    \multirow{4}{*}{{Low-Resource}} 
    & Adapter   & 63.0 & 62.5 & 52.4 & 35.3 & 87.6 & 57.5 & 61.1 & 73.4 & 61.6\\
    & LoRA      & \textbf{68.8} & \textbf{62.7} & \textbf{62.9} & 38.2 & \textbf{89.4} & 56.5 & 64.2 & 74.5 & \textbf{64.7}\\
    & IA3       & 67.0 & 50.3 & 60.6 & \textbf{40.9} & 86.0 & \textbf{59.2} & 64.0 & \textbf{75.4} & 62.9 \\
    & Prefix    & 49.8 & 56.3 & 51.3 & 20.6 & 81.6 & 51.6 & \textbf{65.6} & 53.1 & 53.7\\
    \midrule
    \multirow{4}{*}{{Medium-Resource}} 
    & Adapter   & 74.9 & 63.5 & 72.9 & 66.5 & 85.4 & 58.1 & 64.1 & 73.3 & 69.8\\
    & LoRA      & \textbf{80.0} & \textbf{66.4} & \textbf{78.9} & \textbf{74.8} & 78.3 & 54.7 & 65.2 & 72.3 & 71.3\\
    & IA3       & 77.4 & 55.1 & 77.8 & 76.4 & 88.5 & 59.3 & 65.6 & \textbf{75.0} & \textbf{71.9} \\
    & Prefix    & 56.5 & 52.2 & 63.1 & 38.8 & \textbf{88.9} & \textbf{60.0} & \textbf{65.9} & 74.7 & 62.5\\
    \midrule
    \multirow{4}{*}{{High-Resource}} 
    & Adapter   & 79.8  & \textbf{66.0} & 81.3 & 80.2 & \textbf{91.9} & 59.8 & 64.2 & 73.3 & 74.6\\
    & LoRA      & \textbf{84.7} & 64.5 & \textbf{84.9} & \textbf{87.1} & 83.5 & 55.9 & 65.8 & 72.4 & \textbf{74.9}\\
    & IA3       & 80.9 & 58.3 & 84.0 & 85.0 & 88.9 & 60.5 & 65.8 & 74.7 & 74.8 \\
    & Prefix    & 67.5 & 55.7 & 70.3 & 52.1 & 91.0 & \textbf{61.3} & \textbf{67.6} & \textbf{76.3} & 67.7\\

    \bottomrule
    \end{tabular}%
    \caption{Fine-tuned average results with all PEFT methods on datasets of different sizes.}
    \label{tab:scale}
    \vspace{-0.3cm}
\end{table*}

\vic{In practical applications, MLLMs often require fine-tuning on downstream datasets \cite{li2023LLaVAmed,you2023ferret}, making PEFT an efficient choice. However, the sizes of these specific task datasets may vary, leading to the question: \emph{How to select PEFT methods for datasets of different scales when training?} Therefore, we investigate the performance of PEFT methods on datasets of varying scales. We followed \citet{chen-etal-2022-revisiting} resource setting, and randomly sampled 1k, 5k, and 10k data points from the training set of each dataset.  We categorize 1k data points as \textbf{Low-Resource}, 5k data points as \textbf{Medium-Resource}, and 10k data points as \textbf{High-Resource}. Note that since the training set of OKVQA contains only 9k samples, we considered the full data as high-resource. Table~\ref{tab:scale} presents the results.}


\vic{Our main findings are the  following:} 

\begin{itemize}
\item
\vic{\emph{High-Resource will make the MLLM more powerful, while Medium-Resource will be more efficient.} The performance of the four PEFT methods improves as the scale of resources grows, i.e.\ all achieve their best performance with high-resource. Thus, when resources are sufficient, fine-tuning on high-resource datasets will yield better performance. Average performance improvement is shown in App.~\ref{app:d}.}

\item
\vic{\emph{The unseen datasets tend to favor more resources.} When fine-tuning on an unseen dataset with more data, all PEFT methods show a significant performance improvement. In contrast, as the resources of the dataset increase, we did not observe significant performance improvement on seen datasets.}
\end{itemize}


\subsection{Stability Analysis}


\vic{\citet{he-etal-2021-effectiveness} and \citet{chen-etal-2022-revisiting} carried out experiments with different random seeds to investigate the instability of fine-tuning LLMs using  PEFT methods. Analogously, we look at such instability for MLLMs. We concentrate on the SQA (img) from the unseen datasets and OKVQA from seen datasets and select three random seeds: \textbf{[seed21, seed42, seed63]}. We present a stability analysis for LLaVA-1.5-7B,  more analysis can be found in App.~\ref{app:e}}.
%
%

\vic{The number of trainable parameters plays a crucial role in the fine-tuning process of a model. However, in the multimodal setting, the relationship between the number of trainable parameters and stability when fine-tuning with PEFT is not yet clear.} 
\vic{With this in mind, we look at the following question: \emph{Does fewer trainable parameters lead to higher stability?} We conducted an experiment under different trainable parameter conditions: on seed 21, seed 42, and seed 63, and varied the Lora Rank, Adapter Bottleneck Size, and Virtual Tokens to control the number of trainable parameters. Table~\ref{tab:seed_para} presents the performance of various PEFT methods and their standard deviations under different numbers of trainable parameters. IA3 is not tested since its trainable parameters cannot be modified. We draw the following conclusions: }


\begin{table*}[t]
\scriptsize
    \centering
    \begin{tabular}{cc|lcccc}
    \toprule
     & Source domain & Target domain & overfitting epoch 1 & overfitting epoch 2 & overfitting epoch 3 & overfitting epoch 4\\ 
     
        \midrule
        \multirow{7}{*}{Adapter}
        & \multirow{2}{*}{IconQA-txt} 
        & SQA (img) & 56.0 & \textbf{56.4} & 56.0  & 56.0\\
        & & VizWiz & 58.7 & \textbf{58.9} & \textbf{58.9}  & 58.2\\
        & \multirow{2}{*}{SQA (img)} 
        & IconQA-txt & 36.7 & 36.9 & \textbf{37.3} & \textbf{37.3}\\
        & & VizWiz & \textbf{58.3} & 58.0 & 57.8  & 57.7\\
        & \multirow{2}{*}{VizWiz} 
        & SQA (img) & \textbf{56.9} & 56.7 & 56.2 &56.1 \\
        & & IconQA-txt & \textbf{47.6} & 45.9 & 44.9 & 44.4\\
        & Avg & - & \textbf{52.4} & 52.1 & 51.9 &51.6 \\
        \midrule
        \multirow{7}{*}{LoRA}
        & \multirow{2}{*}{IconQA-txt} 
        & SQA (img)  & 50.8 & 51.8 & \textbf{52.5} & 52.4\\
        & & VizWiz  & \textbf{58.5} & 57.6 & 57.8 & 57.6\\
        & \multirow{2}{*}{SQA (img)} 
        & IconQA-txt & 33.8 & \textbf{33.9} & 33.5 & 33.4\\
        & & VizWiz & 56.8 & \textbf{57.2} & 56.9  & 56.8 \\
        & \multirow{2}{*}{VizWiz} 
        & SQA (img) & \textbf{61.3} & 59.9 & 59.1  & 59.0\\
        & & IconQA-txt & 42.2 & 44.3 & \textbf{46.9} & 40.9\\
        & Avg & - & 50.6 & 50.8 & \textbf{51.1} & 50.0\\
        \midrule
        \multirow{7}{*}{IA3}
        & \multirow{2}{*}{IconQA-txt} 
        & SQA (img) & 61.1 & \textbf{61.5} & 61.0 & 61.2\\
        & & VizWiz & 43.4 & 42.3 & \textbf{45.2} & 43.5\\
        & \multirow{2}{*}{SQA (img)} 
        & IconQA-txt & \textbf{45.2} & 44.7 & 44.0 &43.5\\
        & & VizWiz & \textbf{53.5} & 53.1 & 53.1 &52.9\\
        & \multirow{2}{*}{VizWiz} 
        & SQA (img) & \textbf{61.4} & 60.2 & 60.4 & 60.0\\
        & & IconQA-txt & 42.6 & \textbf{43.3} & 41.1 &41.8\\
        & Avg & - & \textbf{51.2}& 50.9& 50.8& 50.5\\
        \midrule
        \multirow{7}{*}{Prefix}
        & \multirow{2}{*}{IconQA-txt} 
        & SQA (img) & \textbf{41.1} & 37.4 & 34.7 & 33.9\\
        & & VizWiz & \textbf{47.2} & 43.6 & 41.8 & 41.3\\
        & \multirow{2}{*}{SQA (img)} 
        & IconQA-txt & 28.3 & 32.6&\textbf{33.1} & 32.8\\
        & & VizWiz & \textbf{54.0} & 53.7 & 53.9 & 53.5\\
        & \multirow{2}{*}{VizWiz} 
        & SQA (img) & \textbf{47.1} & 39.9 & 41.4 & 46.3\\
        & & IconQA-txt  & 40.3 & \textbf{44.5} & 40.0 &40.6\\
        & Avg & - & \textbf{43.0}& 42.0& 40.8& 41.4\\
    \bottomrule
    \end{tabular}%
   \caption{Performance on target domain with different PEFT methods. For each target domain and PEFT method, four epochs closest to the optimal point of overfitting were selected to test on the target domain. Avg: The average results of target domain at each epoch.}
    \label{tab:generalization_main}
    \vspace{-0.1cm}
\end{table*}

\begin{figure}[h]
    \centering
    \includegraphics[width=0.8\linewidth]{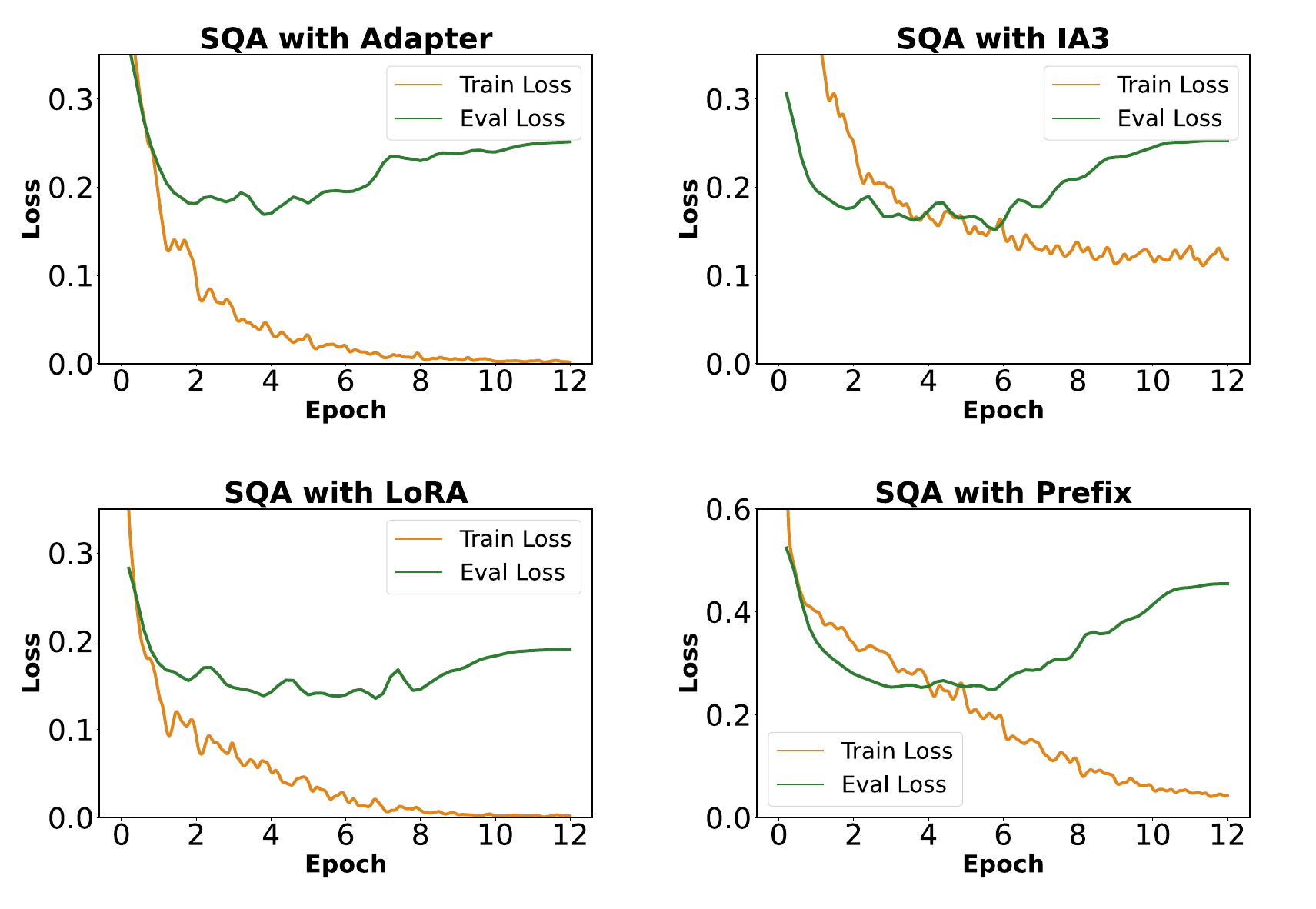}
    \caption{Train-Eval loss of all PEFT methods on SQA (img). The orange line shows Train Loss. Eval loss is colored with green.}
    \label{fig:eval_curve}
\end{figure}

\begin{figure}[h]
    \centering
    \includegraphics[width=0.8\columnwidth]{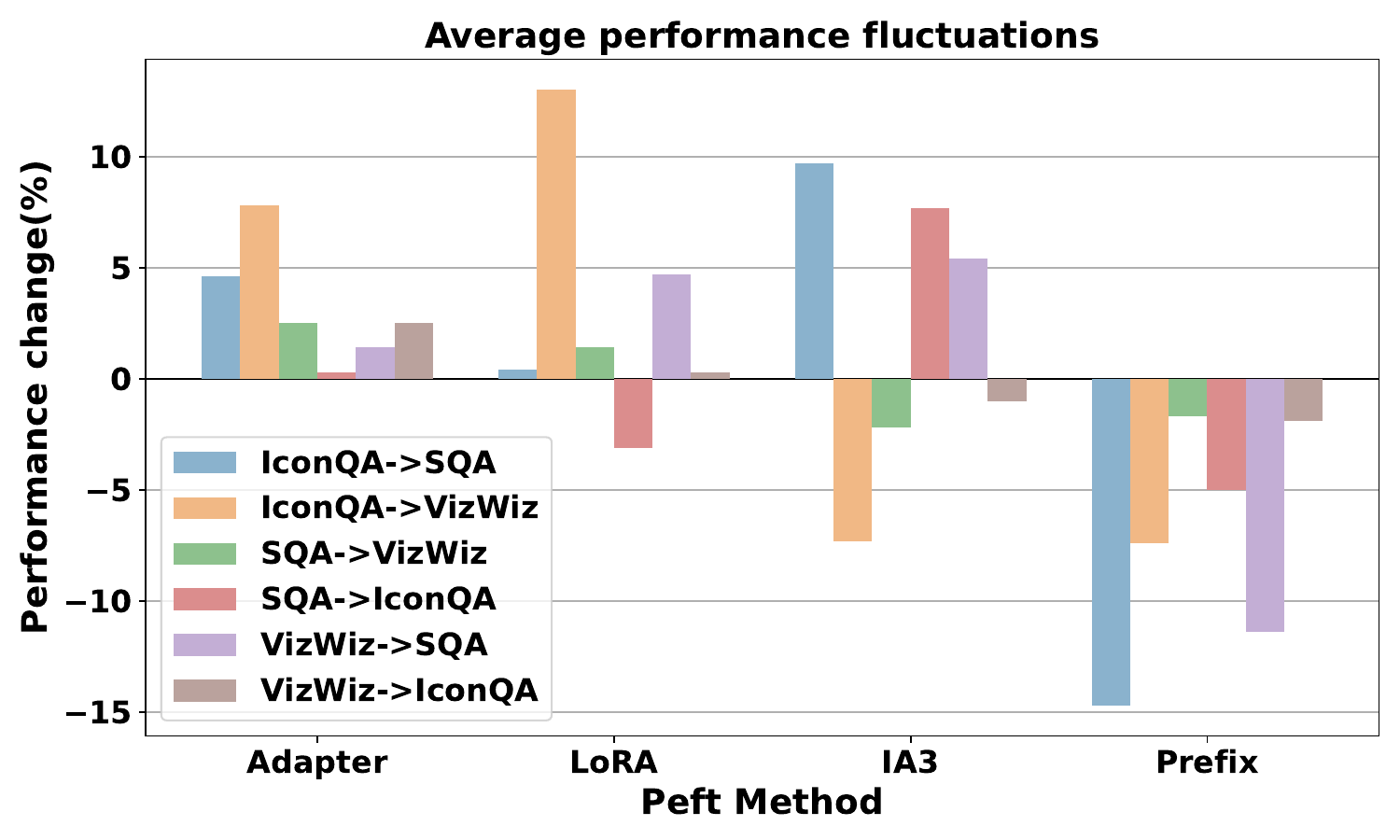}
    \caption{Average performance fluctuation of four epochs on each source-target domain. We calculate the mean of four PEFT methods on each source-target domain and display the average performance fluctuation of all PEFT methods on those domain-pair.}
    \label{fig:fluctuation}
    \vspace{-0.4cm}
\end{figure}

\begin{itemize}[itemsep=0.5ex, leftmargin=5mm]
\setlength{\itemsep}{0pt}
\setlength{\parsep}{0pt}
\setlength{\parskip}{0pt}

\item \vic{   \emph{Adapter and LoRA exhibit drastically different levels of stability on the unseen and seen datasets. Prefix-Tuning shows a strong instability on the unseen datasets.} Adapter gradually stabilizes with decreasing parameters on OKVQA, but becomes unstable with fewer parameters on SQA (img). Conversely, LoRA becomes unstable with decreasing parameters on OKVQA, but stabilizes with fewer parameters on SQA (img). Prefix-Tuning exhibits stability on OKVQA, and shows a relatively stable performance with fewer parameters on SQA (img).}
    
\end{itemize}

\vspace{-3mm}

\subsection{\vi{Overfitting and Generalization}}

\vic{\emph{How robust different PEFT methods are relative to overfitting?} To address this question, we considered three datasets from the unseen datasets: SQA (img), IconQA-txt, and Vizwiz. We choose one dataset from \xt{these three datasets} as the source domain, and fine-tuned LLaVA-1.5-7B with PEFT methods on each source domain for 12 epochs.}

\begin{table*}[th]
    \centering
    \scriptsize
    \resizebox{\textwidth}{!}{%
    \begin{tabular}{c|c|cc|cccccccc}
    \toprule
        Fine-tuning Task & Method & Overall Score↑  & Hallucination Rate↓ & Attribute & Adversarial & Comparison & Counting & Relation & Environment & Holistic & Other \\
    \midrule
    \multirow{4}{*}{IconQA-txt}
        & Adapter & \textbf{1.08} & \textbf{0.70} & 0.58 & 1.17 & 1.42 & 0.25 & 2.17 & 1.33 & 0.00 & 1.75 \\ 
        & LoRA & 0.69 & 0.83 & 0.58 & 0.00 & 1.42 & 0.67 & 1.17 & 1.00 & 0.00 & 0.67 \\ 
        & IA3 & 0.76 & 0.83 & 0.42 & 0.42 & 1.67 & 1.00 & 0.50 & 0.75 & 0.00 & 1.33 \\ 
        & Prefix & 1.00 & \textbf{0.70} & 2.33 & 0.75 & 0.25 & 1.00 & 1.00 & 1.25 & 0.00 & 1.42 \\
    \midrule
    \multirow{4}{*}{Flickr30k}
        & Adapter & \textbf{0.73} & \textbf{0.81} & 0.08 & 0.00 & 1.92 & 0.75 & 0.33 & 1.08 & 0.27 & 0.92 \\
        & LoRA& 0.70 & 0.83 & 0.75 & 0.33 & 1.92 & 0.58 & 0.42 & 0.33 & 0.00 & 1.25 \\
        & IA3 & 0.70 & 0.84 & 0.25 & 0.42 & 1.33 & 1.17 & 0.75 & 0.67 & 0.00 & 1.00 \\
        & Prefix & 0.59 & 0.82 & 0.25 & 0.00 & 0.50 & 0.33 & 0.58 & 1.50 & 0.00 & 1.33 \\
    \bottomrule
    
    \end{tabular}%
    }
    \caption{Evaluation results of LLaVA-1.5-7B with different PEFT methods on MMHAL-Bench. }
    \label{tab:hal_mmhal}
\end{table*}

\vic{\emph{Adapter and LoRA exhibit stronger robustness.} Figure~\ref{fig:eval_curve} (and Figure~\ref{fig:app_over} in App.~\ref{app:f}) shows the evaluation loss of various PEFT methods on SQA as the number of training epochs changes. We observe that when overfitting occurs, there are differences in the robustness exhibited by each PEFT method on each dataset. On SQA, LoRA, IA3, and Adapter a relatively strong robustness is demonstrated, with LoRA performing the best. Prefix-Tuning shows poor robustness on SQA. App.~\ref{app:f} provides further analysis on IconQA-txt and Vizwiz.}



\vic{When facing overfitting an important question is \emph{How do various PEFT methods perform in terms of generalization? And how to achieve the best generalization performance during training?} With these questions in mind, we conducted the next experiment. Based on Figure~\ref{fig:eval_curve}, we identified the training step with the minimum evaluation loss for each PEFT method.  We selected four overfitting points which are the closest to the minimum evaluation loss point on the source domain. Subsequently, we tested the performance of each epoch on the other two target domains, yielding the results shown in Table~\ref{tab:generalization_main}. We draw the following conclusions.}

\begin{table}[t]
\scriptsize
    \centering
    \begin{tabular}{cc|cc}
    \toprule
        &  & OKVQA & SQA (img)\\
    \midrule
    \multirow{4}{*}{Adapter}
         & Bottleneck Size=32   & $62.9_{\pm{0.21}}$  & $80.4_{\pm{3.12}}$  \\
         & Bottleneck Size=64   & $62.7_{\pm{0.60}}$  & $81.2_{\pm{2.75}}$  \\
         & Bottleneck Size=128  & $61.4_{\pm{0.20}}$  & $81.3_{\pm{1.80}}$  \\
         & Bottleneck Size=256  & $58.8_{\pm{0.84}}$  & $82.2_{\pm{1.91}}$  \\
    \midrule
    \multirow{4}{*}{LoRA}
         & LoRA Rank=16   & $56.1_{\pm{0.53}}$  & $85.7_{\pm{0.32}}$   \\
         & LoRA Rank=32   & $56.1_{\pm{0.27}}$  & $85.3_{\pm{0.92}}$   \\
         & LoRA Rank=64   & $56.4_{\pm{0.20}}$  & $85.0_{\pm{0.85}}$   \\
         & LoRA Rank=128  & $56.6_{\pm{0.12}}$  & $85.4_{\pm{0.85}}$   \\
    \midrule
    \multirow{4}{*}{Prefix}
         & Virtual Tokens=10     & $62.2_{\pm{0.10}}$  & $73.4_{\pm{2.62}}$  \\
         & Virtual Tokens=20     & $61.5_{\pm{0.20}}$  & $72.2_{\pm{1.11}}$  \\
         & Virtual Tokens=30     & $61.2_{\pm{0.06}}$  & $67.7_{\pm{0.78}}$  \\
         & Virtual Tokens=40     & $61.2_{\pm{0.27}}$  & $56.2_{\pm{19.20}}$ \\
    \bottomrule
    \end{tabular}%
    \caption{Performance on three PEFT methods with different hyperparameter settings. Reported results are averages across three runs with different random seeds.}
    \label{tab:seed_para}
\end{table}

\begin{itemize}[itemsep=0.5ex, leftmargin=5mm]
\setlength{\itemsep}{0pt}
\setlength{\parsep}{0pt}
\setlength{\parskip}{0pt}
    \item \vic{\emph{Adapter exhibits  the \xt{strongest} in generalization.}  Figure~\ref{fig:fluctuation} shows that when a model is fine-tuned using Prefix-Tuning its generalization performance is quite poor. Models using Adapter consistently exhibit a good generalization performance regardless of the situation, while the generalization performance of a model fine-tuned with Prefix-Tuning is consistently negative. Models using LoRA and IA3 show fluctuation in generalization performance.}

    \item \vic{ \emph{IA3, Adapter, and Prefix-Tuning show the best generalization performance at the \xt{first overfitting} epoch}. From the results of Table~\ref{tab:generalization_main}, we can find that Adapters, IA3, and Prefix-Tuning, all achieve the best average generalization performance at the first overfitting epoch. In general, the model's generalization weakens as overfitting intensifies. However, LoRA achieves the best model generalization performance at the third overfitting epoch, indicating that models fine-tuned with LoRA exhibit the best generalization when overfitting reaches a certain level, gradually weakening afterwards.}
\end{itemize}

\subsection{Hallucination}

\vic{The hallucination problem in LLMs has been widely acknowledged \cite{10.1145/3571730,gudibande2024the}. Since MLLMs are built upon LLMs, this problem is also present in them. \citet{zhai2023investigating} found that further fine-tuning with multimodal instruction-following data leads to hallucinations. Therefore, we aim to investigate the following question: \emph{Which PEFT method results in fewer hallucinations during fine-tuning?} We select IconQA-txt as the source domain and the Flickr30k dataset as the target domain to assess the out-of-domain hallucinations of models fine-tuned with various PEFT methods. App.~\ref{app:h_1} elaborates on how we evaluated the model's hallucination.}

\begin{table}[t]
\scriptsize
    \centering
    \begin{tabular}{c|ccccc}
    \toprule
         Method & Epoch 3 & Epoch 6 & Epoch 9 & Epoch 12 & Avg \\
    \midrule     
         Adapter& 17 & 12 & 14 & 10 & 13.3 \\
         LoRA   & 15 & 14 & 18 & 20 & 16.8 \\
         IA3    & 14 & 17 & 17 & 16 & 16.0 \\
         Prefix & 24 & 18 & 27 & 31 & 25.0 \\
    \bottomrule
    \end{tabular}%
    \caption{Hallucinations statistic of PEFT methods on four epochs. We selected 100 hallucination-free examples from 1k random sampled data from the outputs of LLaVA-1.5-7B. We examined the outputs of LLaVA-1.5-7B with four PEFT methods on those examples, table presents the number of outputs with hallucination.}
    \label{tab:hal_number}
\end{table}

\vic{MMHAL-Bench~\cite{2023llavarlhf} is used to evaluate the hallucinations induced by fine-tuning LLaVA-1.5-7B with four PEFT methods on Flickr30k and IconQA-txt, yielding the results presented in Table~\ref{tab:hal_mmhal}. The results show that Adapter consistently achieved the highest Avg Score and the lowest Hallucination Rate across both fine-tuning tasks. This is consistent with our manual evaluation results.}

\vic{\emph{Adapter demonstrates  potential for addressing hallucinations in MLLMs.} The results are illustrated in Table~\ref{tab:hal_number}. We observe that Adapter achieves the lowest average hallucination rate across four epochs, at only 13.3\%. It can also be found that other PEFT methods tend to produce more hallucinations after further fine-tuning, especially Prefix-Tuning, which generates an additional 24\% of hallucinations from epoch 3 to epoch 12. In contrast, with further fine-tuning, Adapter reduced the number of hallucinations produced. In line with previous studies~\cite{wang2023cogvlm}, we attribute this phenomenon to the new parameters in the Adapter method, which provides a new module to adapt to downstream datasets while keeping the base model's original weights.}

\section{Conclusion}

\vic{We conducted an extensive investigation on four PEFT methods applied to MLLMs across different multimodal tasks. By fine-tuning different MLLMs in a uniform way and conducting thorough hyper-parameter optimization, we benchmarked the performance of these methods. Our findings indicate that Adapter excels in accuracy, stability, generalization, and producing fewer hallucinations. Additionally, we found that fine-tuning the connector layers of MLLMs simultaneously does not always yield better results. Finally, comprehensive ablation studies were performed to understand the contributions of the location of PEFT modules, learning rate settings, and the size of training data on PEFT performance.}

\section*{Limitations}
\vic{All our experiments were conducted within the defined framework, which involves connector layers serving as the bridge between the visual encoder and LLM, and no additional modules were inserted on the LLM. Due to the limitation of computational resources, we have currently employed only a subset of datasets to conduct our analysis. Additionally, our choice of MLLMs on the analysis experiments is limited to LLaVA-1.5-7B or Qwen-VL-Chat. In the future, we plan to conduct an analysis on more datasets and MLLMs. }

\bibliography{anthology}

\clearpage
\appendix

\section{Datasets Setup}
\label{appendix_A}

\begin{table*}[!htp]
\small
    \centering
    \resizebox{\textwidth}{!}{%
    \begin{tabular}{l|l|l|l|l|p{5cm}|l|l|l}
    \toprule
         Dataset & Task & Split & Metric & Answer type & Description & Dataset Type & \# Train & \# Test (Val)\\
    \midrule
         Flickr30K & Image Caption & train \& test & CIDEr ($\uparrow$) &  Caption & Image dataset with captions for natural scenes. & Unseen & 31k &4k\\
         \midrule
         IconQA-blank & Visual Reasoning & train \& test & Accuracy ($\uparrow$) & Word & Visual reasoning with abstract icons, no text. & Unseen & 11k &4k\\
         \midrule
         IconQA-txt & Visual Reasoning & train \& test & Accuracy ($\uparrow$) & Word & Abstract icon reasoning with textual hints.& Unseen & 19k & 6k\\
         \midrule
         OKVQA & Knowledge Grounded VQA & train \& test & VQA-Score ($\uparrow$) & Phrase & VQA requiring external knowledge. & Seen & 9k & 5k\\
         \midrule
         SQA(img) & Knowledge Grounded VQA & train \& test & Accuracy ($\uparrow$) & Option & Science-focused multiple-choice VQA &Unseen & 13k & 4k\\
         \midrule
         OCRVQA & Reading Comprehension VQA & train \& test & Accuracy ($\uparrow$) & Phrase & VQA with text recognition in images.&seen & 20k & 5k\\
         \midrule
         VQAv2 & General VQA & train \& test & VQA-Score ($\uparrow$) & Phrase & Diverse open-ended visual question answering. &Unseen & 20k & 5k\\
         \midrule
         VizWiz & General VQA & train \& val & VQA-Score ($\uparrow$) & Phrase & VQA sourced from visually impaired users' photos.&seen & 20k & 4k\\
    \bottomrule
    \end{tabular}%
    }
    \caption{Detailed description for the datasets we used, including task types, training and test split, evaluation metric, statistic, dataset type, and the type of answer. To be specific, ``Seen'' means that the dataset has been used as a pre-training dataset in the model being evaluated. ``Unseen'' refers to datasets that have not been encountered by the model.}
    \label{tab:app_dd}
\end{table*}

\subsection{Detailed Description of Datasets}
\label{appendix_A_1}


Since OCRVQA and VQAv2 are very large, we randomly extract 20k samples from their training sets as the new training sets and another 5k samples from the test sets to create the new test sets. We utilized a variety of multimodal datasets for fine-tuning and evaluating. Detailed information for each dataset is provided in Table~\ref{tab:app_dd} below.

\subsection{Seen Datasets for All MLLMs}
\label{appendix_A_2}

\begin{table*}[htb]
    \centering
    \footnotesize
    \begin{tabular}{ll|p{10cm}}
    \toprule
        model & Phrase & Seen datasets \\
        \midrule
        \multirow{2}{*}{LLaVA} 
        & \multirow{1}{*}{Pretrained} & CC-595K, LLaVA-Instruct-158K \\
        & \multirow{1}{*}{Fine-tuning} & \textbf{VQAv2}, \textbf{OKVQA}, \textbf{OCRVQA}, GQA, A-OKVQA, TextCaps, RefCOCO, VG \\
        \midrule
        \multirow{5}{*}{Qwen-VL-Chat} 
        & \multirow{4}{*}{Pretrained} & LAION-en, LAION-COCO, DataComp, Coyo, CC12M, CC3M, SBU, COCO Caption, LAION-zh, GQA, VGQA, VQAv2, DVQA, OCRVQA, DocVQA. TextVQA, ChartQA, AI2D, GRIT, VG, RefCOCO+, RefCOCOg, SynthDoG-en \& zh, Common Crawl of pdf \& HTML, In-house Data \\
        & \multirow{1}{*}{Fine-tuning} & \textbf{OKVQA}, \textbf{OCRVQA}, \textbf{VQAv2} \\
        \midrule
        \multirow{3}{*}{ShareGPT4v} 
        & \multirow{1}{*}{Pretrained} & CC-595K, LLaVA-Instruct-158K, ShareGPT4V-PT \\
        & \multirow{2}{*}{Fine-tuning} & \textbf{VQAv2}, \textbf{OKVQA}, \textbf{OCRVQA}, A-OKVQA, GQA,  TextCaps, RefCOCO, VG, ShareGPT4V \\

    \bottomrule
    \end{tabular}%
    \caption{The datasets used during the pretraining and further fine-tuning processes of MLLMs.}
    \label{tab:app_sd}
\end{table*}

In our experiments, we divided the datasets into unseen datasets and seen datasets. In Table~\ref{tab:app_sd}, we present the training datasets used by each MLLM in our experiments. It is worth noting that, due to the data used in both pre-training and fine-tuning stages being filtered or sampled from the datasets,  to ensure a fair comparison, it is imperative that each seen dataset is fully seen by the model. We composed a mixed dataset consisting of three seen datasets: OKVQA, VQAv2, and OCRVQA. Subsequently, we kept the visual encoder frozen and conducted a Full Fine-Tuning on each model using this mixed dataset. During this process, we set the learning rate to 2e-5 and the global batch size to 128. We maintained all other settings the same as those used in the original paper for Full Fine-Tuning~\cite{liu2023improved,bai2023qwenvl}.

\subsection{Instruction-following Data Template}
\label{appendix_a_3}

\begin{table*}[htb]
    \centering
    \resizebox{0.95\textwidth}{!}{%
    \begin{tabular}{l|p{14cm}}
    \toprule
    \textbf{Model} & Image Annotation[IMAGE] \\
    \midrule
    LLaVA Image & <Image> \\
    Qwen Image & <img><$/$img> \\
    
    \midrule
         \textbf{Task} & Instruction Template \\
    \midrule
         \multirow{9}{*}{Image Caption}
         & [IMAGE] Share a concise interpretation of the image provided.\\
         & [IMAGE] Render a clear and concise summary of the photo.\\
         & [IMAGE] Write a terse but informative summary of the picture.\\
         & [IMAGE] Offer a succinct explanation of the picture presented.\\
         & [IMAGE] Describe the image concisely.\\
         & [IMAGE] Provide a brief description of the given image.\\
         & [IMAGE] Create a compact narrative representing the image presented. \\
         & [IMAGE] Relay a brief, clear account of the picture shown.\\
         & [IMAGE] Summarize the visual content of the image. \\
         \midrule
         & [IMAGE] Give a short and clear explanation of the subsequent image \\
         \multirow{1}{*}{Knowledge Grounded VQA} 
         & [IMAGE] \{Question\} Answer the question using a single word or phrase.\\
         & [IMAGE] \{Question\} A.choice 1, B.choice 2, C.choice 3, ...\\
         \midrule
         \multirow{3}{*}{Visual Reasoning} 
         &  [IMAGE] \{Question\} Fill in the blanks in (\_) or answer this question.\\
         & [IMAGE] \{Question\} Choices: choice 1, choice 2, choice 3,... Choose an option from the choices to answer the question.\\
         \midrule
         \multirow{1}{*}{Reading Comprehension VQA} 
         & [IMAGE] \{Question\} Answer the question using a single word or phrase.\\
         \midrule
         \multirow{4}{*}{General VQA} 
         & [IMAGE] Question: \{Question\}\\
         & [IMAGE] Question: \{Question\} When the information is insufficient, respond with "Unanswerable". Answer the question using a single word or phrase. \\
         & [IMAGE] \{Question\} Answer the question using a single word or phrase.\\
    \bottomrule
    \end{tabular}%
    }
    \caption{The instruction format of different tasks when we fine-tune MLLMs.}
    \label{tab:app_template}
\end{table*}

Multimodal Instruction-Following Tuning is a crucial component for the success of MLLMs. The MLLMs used in this experiment follow the same approach in data processing as the original models. Therefore, there are slight differences in the processing of LLaVA-1.5, ShareGPT4v, and Qwen-VL-Chat. They employ different image annotations, but the data instruction format remains consistent. Table~\ref{tab:app_template} shows the template of all dataset types used in the experiments.

\section{Models and Hyperparameters}
\label{appendix_B}

\begin{figure*}[ht]
  \centering
  \begin{subfigure}[b]{0.32\textwidth}
    \centering
    \includegraphics[width=\textwidth]{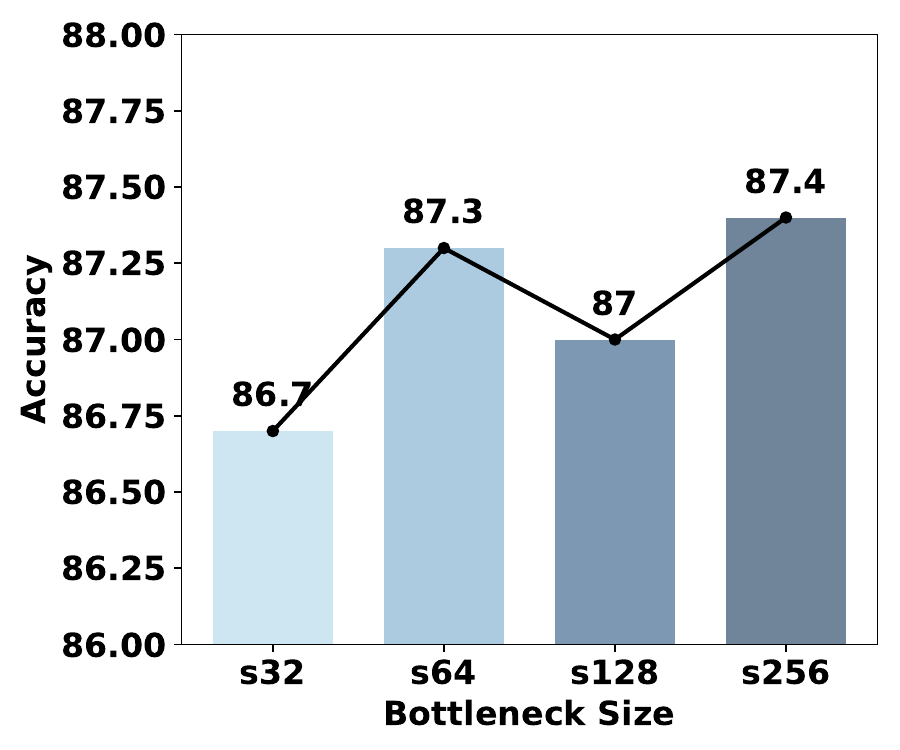}
    \caption{Adapter}
    \label{fig:param_adapter}
  \end{subfigure}
  \begin{subfigure}[b]{0.32\textwidth}
    \centering
    \includegraphics[width=\textwidth]{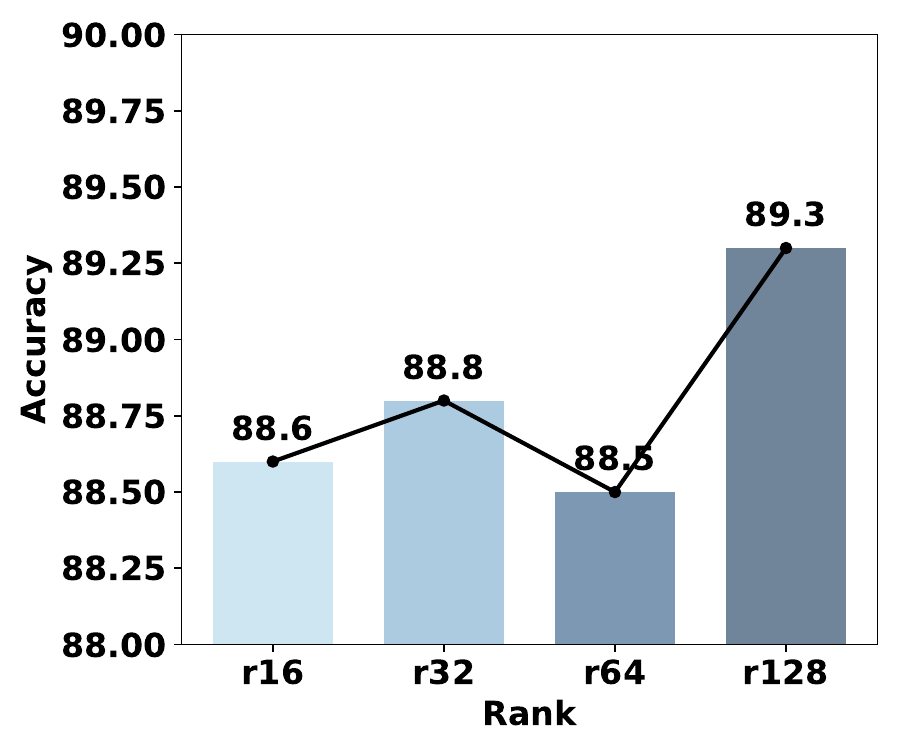}
    \caption{LoRA}
    \label{fig:param_lora}
  \end{subfigure}
  \begin{subfigure}[b]{0.32\textwidth}
    \centering
    \includegraphics[width=\textwidth]{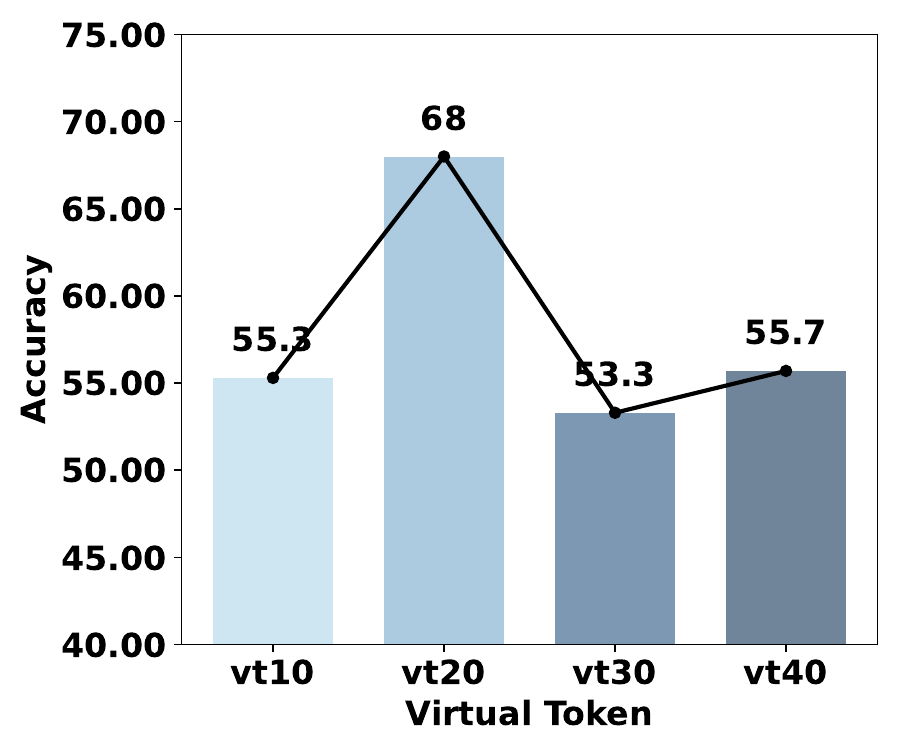}
    \caption{Prefix}
    \label{fig:param_prefix}
  \end{subfigure}
  \caption{Average accuracy of Results of various PEFT parameters on SQA (all). s: Bottleneck Size. r: LoRA Rank. vt: Virtual Token}\label{fig:para_select}
\end{figure*}

\subsection{Models}
\label{app:b_1}

LLaVA-1.5 consists of CLIP-ViT-L/14 \cite{ilharco_gabriel_2021_5143773}, Vicuna-v1.5 \cite{vicuna2023}, and an MLP serving as the connector layer. During fine-tuning with multimodal instruction-following data, LLaVA-1.5 updates only the parameters of the MLP connector and the LLM, while keeping the parameters of the visual encoder frozen. ShareGPTv4 is obtained by fine-tuning  LLaVA-1.5 using the ShareGPT4v dataset \cite{chen2023sharegpt4v}, which comprises 100K high-quality captions from various images generated by GPT4-vision. Qwen-VL-Chat comprises ViT-G/16 \cite{ilharco_gabriel_2021_5143773}, Qwen-7B \cite{bai2023qwen}, and a cross-attention module serving as the connector. During its vision instruction tuning phase, Qwen-VL-Chat updates only the parameters of the connector and the LLM.

\subsection{HyperParameters}
\label{app:b_2}

Following \citet{hu-etal-2023-llm}, we conducted the following parameter selection experiments.  Due to computational constraints, we concentrate on the SQA (all) dataset, which has a test set that contains both multimodal and text-only data. So, our goal is to enhance the model's multimodal performance while maximizing its performance on text-only datasets. 

We choose LLaVA-1.5-7B as the base model set the random seed to 42, freeze the visual encoder, and fine-tune the connector layers and LLM. Note that we do not consider IA3 in this experiment as it cannot change the number of trainable parameters. We look at different parameter settings:  LoRA rank of \{16, 32, 64, 128\}, Adapter bottleneck size of \{32, 64, 128, 256\}, and virtual tokens of \{10, 20, 30, 40\} in Prefix-Tuning.  Figure~\ref{fig:para_select} shows the results with various parameter settings on SQA (all). We observe that for LoRA,  a rank of 128 yielded an accuracy of 89.3\%. Setting the adapter's bottleneck size to 256 resulted in an accuracy of 87.4\%. In the case of Prefix-Tuning, setting the virtual tokens to 20 achieved the best accuracy of 68.0\%.

Based on our findings, for all remaining experiments, we use the following PEFT parameters: \textbf{LoRA Rank=128}, \textbf{Adapter Bottleneck Size=256}, and \textbf{Prefix Virtual Token=20}. More detailed hyperparameter settings are presented in Table~\ref{tab:app_hyperpara}. We utilized two NVIDIA A100 80GB GPUs and DeepSpeed for distributed training.

\begin{table*}[htb]
    \centering
    \footnotesize
    \begin{tabular}{lcccc}
    \toprule
         Configuration & LLaVA-7B & LLaVA-13B & Qwen-VL-Chat & ShareGPT4v \\
    \midrule
         ViT & Vicuna-v1.5-7B & Vicuna-v1.5-7B & Qwen-7B & Vicuna-v1.5-7B \\
         LLM & CLIP-ViT-L/14 & CLIP-ViT-L/14 & ViT-G/16 & CLIP-ViT-L/14 \\
         Connector & MLP & MLP & CrossAttn & MLP \\
         Optimizer & \multicolumn{4}{c}{AdamW} \\
         Connector learning rate & 2e-5 & 2e-5 & 1e-5 & 2e-5 \\
         Learning rate schedule &  \multicolumn{4}{c}{cosine decay} \\
         Warm-up ratio & 0.03 & 0.03 & 0.01 & 0.03 \\
         Weight decay & 0.0 & 0.0 & 0.1 & 0.0 \\
         Global batch size & 128 & 128 & 128 & 128 \\
         Gradient Acc & 1 & 1 & 1 & 1\\
         Training epoch &  \multicolumn{4}{c}{3} \\
         Numerical precision & \multicolumn{4}{c}{bfloat16} \\
    \bottomrule
    \end{tabular}%
    \caption{Training hyperparameters when we use PEFT methods to fine-tune those models.}
    \label{tab:app_hyperpara}
\end{table*}

\section{Training of Visual Encoder Analysis}
\label{app:add}
Qwen-vl-chat unfroze the visual encoder during pre-training, which improved the model's performance. However, most current MLLMs maintain the visual encoder frozen during task-specific fine-tuning. We fine-tuned the visual encoder on SQA and VizWiz for unseen tasks, and on OKVQA and OCRVQA for seen tasks. The results are shown in the Table~\ref{tab:vit}. We found that although unfreezing the visual encoder does not significantly increase training resource consumption, the improvement in model performance is limited and, in most cases, can even lead to performance degradation. Therefore, in our experiments, we adhered to the mainstream setting and kept the visual encoder frozen.

\begin{table*}[htb]
\footnotesize
    \centering
    \begin{tabular}{l|l|ccccc}
        \toprule
        Model & Method & SQA (img) & VizWiz & OKVQA & OCRVQA & Avg \\
        \midrule
        \multirow{8}{*}{LLaVA-1.5-7B}
        & Adapter & 84.4 & \textbf{67.6} & 59.8 & 65.2 & \textbf{69.3} \\
        & -w/ Visual Encoder & 84.2 & 67.1 & 60.5 & 65.1 & 69.2 \\
        & LoRA & \textbf{86.2} & 66.5 & 56.5 & 66.7 & 69.0 \\
        & -w/ Visual Encoder & 85.9 & 66.7 & 55.9 & 66.8 & 68.8 \\
        & IA3 & 82.7 & 61.9 & 60.5 & 67.1 & 68.1 \\
        & -w/ Visual Encoder & 83.4 & 62.2 & 60.3 & 66.8 & 68.2 \\
        & Prefix & 68.2 & 60.8 & \textbf{61.3} & 68.5 & 64.7 \\
        & -w/ Visual Encoder & 65.9 & 62.1 & 60.8 & \textbf{68.7} & 64.4 \\
        \bottomrule
    \end{tabular}
    \caption{Results of tuning visual encoder on LLaVA-1.5-7B with four PEFT methods. w/ Visual Encoder: Tuning the Visual Encoder.}
    \label{tab:vit}
\end{table*}

\section{Location Analysis}
\label{app:C}
We also conducted the Module Location experiment on Qwen-VL-Chat. Table~\ref{tab:qwen_locate} shows the results. We observe that for LoRA and IA3, Both settings achieved the best results. As for Adapter, inserting it only into the MLP layer yielded the best performance. It reveals that the results on Qwen-VL-Chat are consistent with those on LLaVA-1.5-7B.

\begin{table*}[htb]
\footnotesize
    \centering
    \begin{tabular}{llccccccccc}
    \toprule
          Method  & Location & SQA (img) & VizWiz & IconQA-txt & IconQA-blank & OKVQA & OCRVQA & VQAv2 & Avg \\
         \midrule
         \multicolumn{10}{c}{Qwen-VL-Chat} \\
         \midrule
         \multirow{3}{*}{Adapter} 
         & Attn & 85.2  & 64.5  & \textbf{92.2}  & 88.3  & 49.3  & \textbf{71.7}  & 63.9  & 73.6  \\
         & MLP  & 81.2  & 69.3  & 90.8  & 87.5  & 51.1  & 69.3  & 70.7  & \textbf{74.3}  \\
         & Both & \textbf{85.6}  & 67.1  & 91.8  & \textbf{90.5}  & 50.7  & 55.9  & 69.2  & 73.0  \\       
         \midrule
         \multirow{3}{*}{LoRA} 
         & Attn & 80.2  & 67.3  & 80.5  & 87.1  & 43.3  & 69.6  & 39.1  & 66.7  \\
         & MLP  & 74.5  & 60.7  & 78.5  & 88.5  & 43.6  & 67.8  & 60.1  & 67.7  \\
         & Both & 84.0  & 68.8  & 71.9  & 83.5  & 43.5  & 67.0  & 63.3  & \textbf{68.9}  \\
         \midrule
         \multirow{3}{*}{IA3} 
         & Attn & 66.4  & 69.1  & 58.2  & 21.7  & 50.8  & 63.3  & 77.3  & 58.1  \\
         & MLP  & 62.8  & 68.3  & 59.8  & 23.1  & \textbf{51.3} & 62.7  & \textbf{77.6}  & 57.9  \\
         & Both & 67.3  & \textbf{69.8}  & 57.3  & 28.7  & 50.5  & 62.1  & 77.5  & \textbf{59.0}  \\
    \bottomrule
    \end{tabular}%
    \caption{Average results of PEFT module location on Qwen-VL-Chat.}
    \label{tab:qwen_locate}
\end{table*}

\section{Data Scale Analysis}
\label{app:d}

\begin{figure}[htb]
    \centering
    \includegraphics[width=0.9\columnwidth]{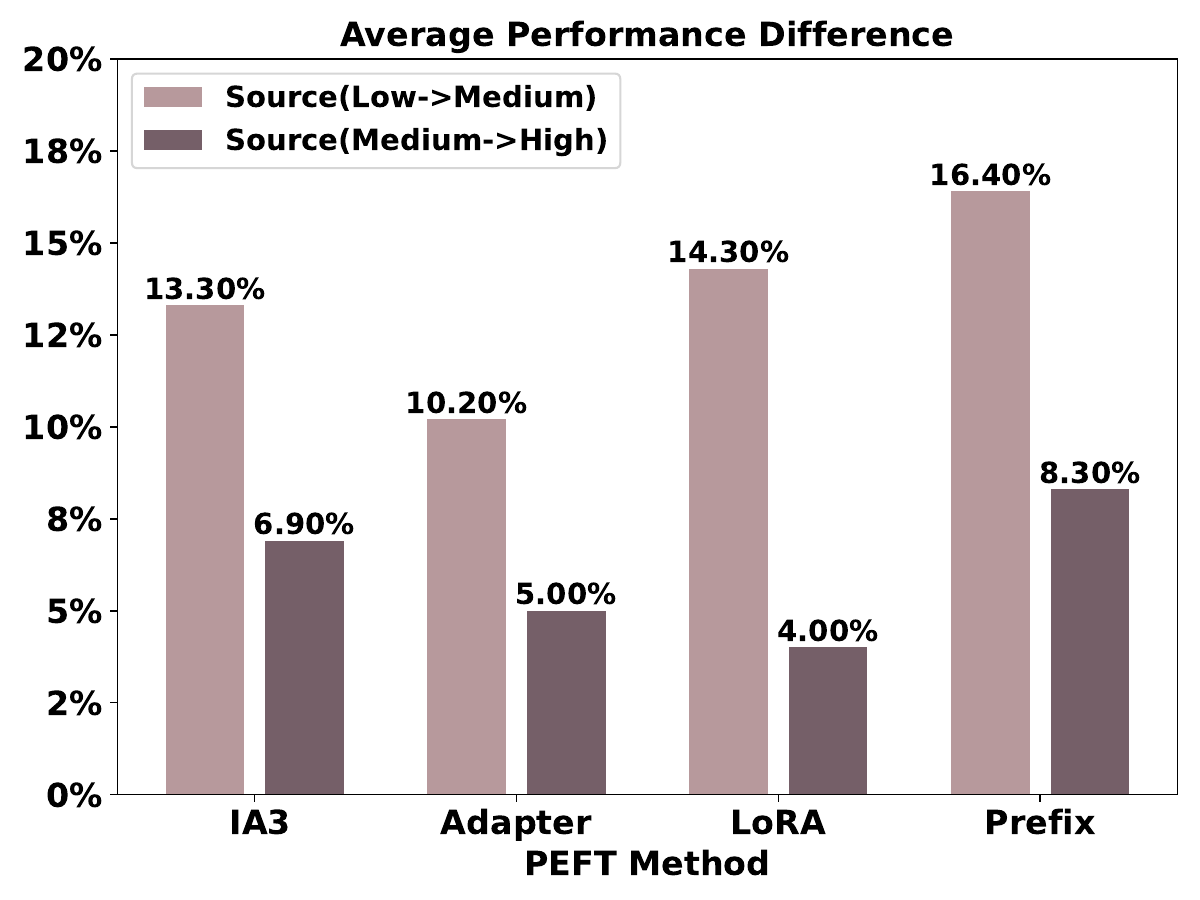}
    \caption{Average performance difference for different PEFT methods in various data-scaling settings. Source (Low->Medium): fine-tuned dataset scaling change from low-resource to high-resource. Source (Medium->High): fine-tuned dataset scaling change from medium-resource to high-resource.}
    \label{fig:app_scale}
    \vspace{-0.5cm}
\end{figure}

Figure~\ref{fig:app_scale} shows the improvement in the average performance of the four PEFT methods as resources transition from low to high. When datasets transition from low to medium resources, all four PEFT methods achieve performance improvements of over 10\%, higher than from medium to high resources. Thus, when computational resources are limited, fine-tuning on medium-resource datasets is more efficient.


\section{Stability Analysis}
\label{app:e}

\begin{table}[ht]
   \resizebox{0.95\columnwidth}{!}{%
    \centering
    \begin{tabular}{cccc}
    \toprule
        &  & OKVQA & SQA (img)\\
    \midrule
    \multirow{4}{*}{Adapter}
         & learning rate=2e-4   & $54.4_{\pm{0.44}}$  & $81.2_{\pm{1.27}}$  \\
         & learning rate=5e-5  & $58.9_{\pm{0.81}}$  & $81.3_{\pm{1.85}}$  \\
         & learning rate=1e-5   & $60.3_{\pm{0.10}}$  & $72.4_{\pm{0.25}}$  \\
         & learning rate=5e-6  & $58.5_{\pm{0.35}}$  & $82.2_{\pm{1.31}}$  \\
    \midrule
    \multirow{4}{*}{LoRA}
         & learning rate=2e-4   & $56.7_{\pm{0.40}} $ &$ 86.0_{\pm{0.35}} $ \\
         & learning rate=5e-5  & $59.8_{\pm{0.10}}$  & $83.1_{\pm{0.45}}$  \\
         & learning rate=1e-5   &$ 62.9_{\pm{0.06}}$  & $74.0_{\pm{2.71}} $ \\
         & learning rate=5e-6  & $61.6_{\pm{0.06}}$  &$ 71.2_{\pm{1.64}}$  \\
    \midrule
    \multirow{4}{*}{IA3}
         & learning rate=2e-4   & $62.7_{\pm{0.98}}$  & $81.7_{\pm{0.87}}$  \\
         & learning rate=5e-5  & $61.4_{\pm{4.89}}$  & $77.1_{\pm{2.54}}$  \\
         & learning rate=1e-5   & $62.9_{\pm{0.20}}$  & $72.2_{\pm{1.28}}$  \\
         & learning rate=5e-6  & $58.8_{\pm{0.06}}$  & $70.8_{\pm{0.64}}$  \\
    \midrule
    \multirow{4}{*}{Prefix}
         & learning rate=2e-4   & $60.9_{\pm{0.21}}$  & $35.3_{\pm{0.85}}$  \\
         & learning rate=5e-5  & $59.6_{\pm{0.06}}$  & $64.6_{\pm{0.01}}$  \\
         & learning rate=1e-5   & $61.5_{\pm{0.06}}$  & $72.3_{\pm{0.21}}$  \\
         & learning rate=5e-6  & $60.8_{\pm{0.01}}$  & $70.2_{\pm{0.01}}$  \\
    \bottomrule
    \end{tabular}%
    }
    \caption{Performance on OKVQA and SQA (img) datasets with different training learning rate of PEFT module.}
    \label{tab:seed_lr}
\end{table}

\begin{figure*}[ht]
  \centering
  \begin{subfigure}{0.49\textwidth}
    \centering
    \includegraphics[width=\textwidth]{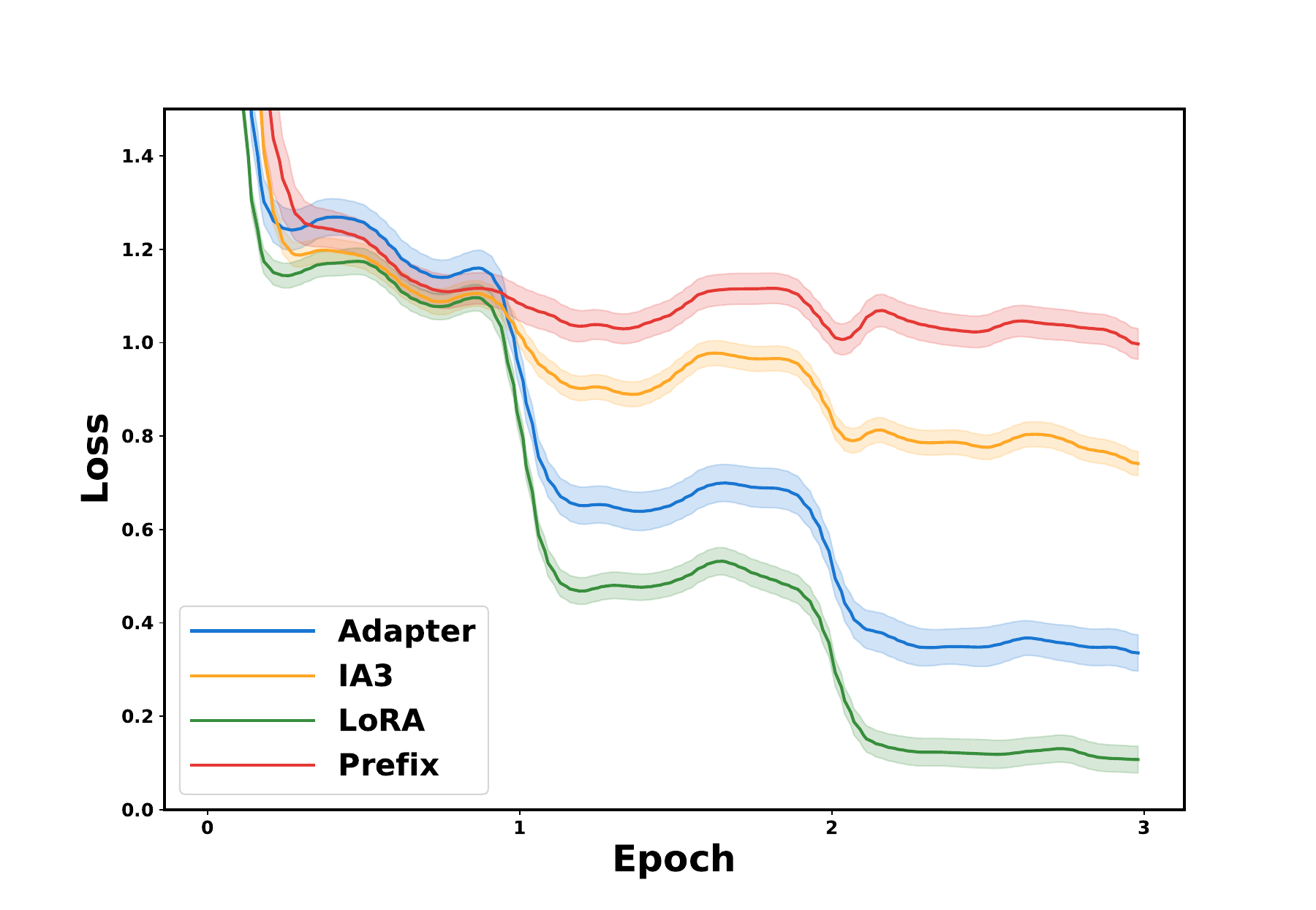}
    \caption{Train Loss of OKVQA}
    \label{fig:train_loss_okvqa}
  \end{subfigure}
  \hfill
  \begin{subfigure}{0.49\textwidth}
    \centering
    \includegraphics[width=\textwidth]{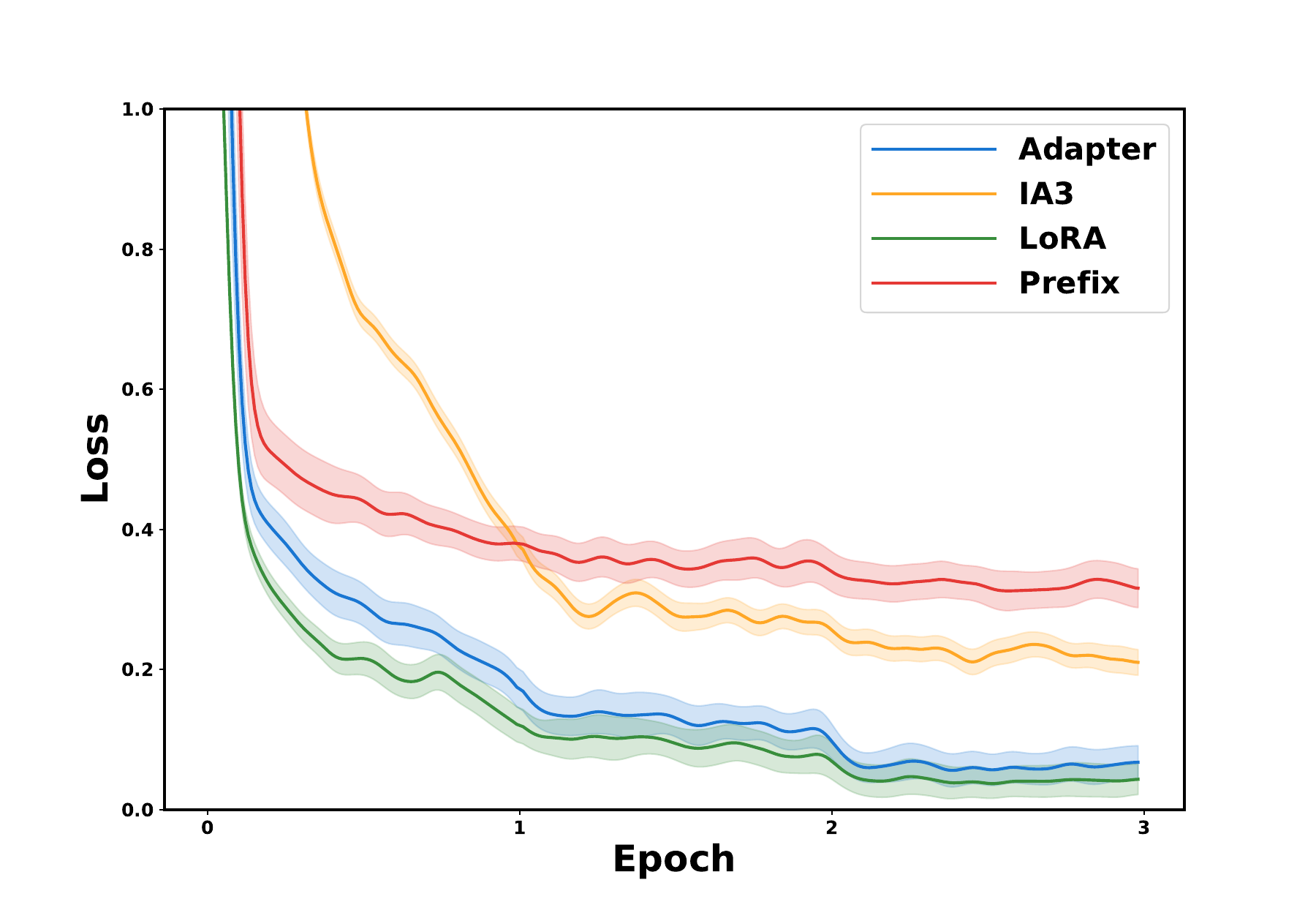}
    \caption{Train Loss of SQA}
    \label{fig:train_loss_sqa}
  \end{subfigure}
  \caption{Train loss reported across three runs with different random seeds. The line plotted the mean of three seeds, where the shaded region represents its 95\% confidence interval.}
  \label{fig:seed_loss}
\end{figure*}

Figure~\ref{fig:seed_loss} presents the training loss, showing that the stability of PEFT varies across different datasets. We observe that Prefix-Tuning and Adapter exhibit larger fluctuations in training loss at each step when trained with different seeds, followed by LoRA, while IA3 shows relatively smaller fluctuations.

We also investigate whether the learning rate correlates with stability. We conducted experiments with learning rates of \{2e-4, 5e-5, 1e-5, 5e-6\}. The results are shown in  Table~\ref{tab:seed_lr}. It can be observed that IA3 and Prefix-Tuning demonstrate more stable performance at smaller learning rates, Prefix-Tuning tends to stabilize gradually as the learning rate decreases.

\section{Overftitting and Generalization Analysis}
\label{app:f}

\begin{figure*}[ht]
    \centering
    \includegraphics[width=1\textwidth]{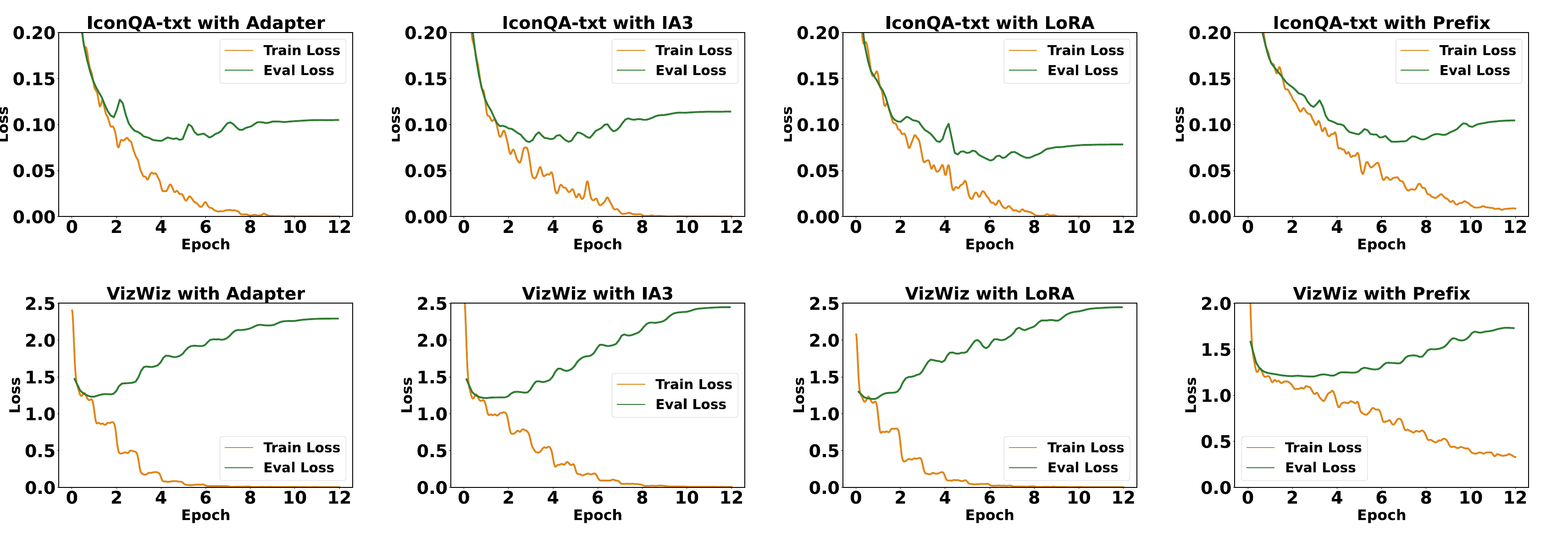}
    \caption{Train-Loss curve on IconQA-txt and Vizwiz.  The orange line shows Train Loss. Eval loss is colored with green.}
    \label{fig:app_over}
\end{figure*}



Overfitting and Generalization experiments were conducted on three unseen VQA datasets. The train-loss curves for IconQA-txt and VizWiz are depicted in Figure~\ref{fig:app_over}. For IconQA-txt, all four PEFT methods exhibit strong robustness, with LoRA performing the best. On Vizwiz, Prefix-Tuning shows the strongest robustness compared to the other three PEFT methods.

\section{Efficiency}
\label{app:g}

We investigate the number of trainable parameters, the training and inference Flops for various MLLMs when fine-tuned with different PEFT methods in this section. Table~\ref{tab:efficiency} shows the results. We derive the training and inference FLOPs in accordance with the methodology outlined in \citet{kaplan2020scaling}. Our analysis shows that models perform better when the connector is not frozen, which suggests that having more trainable parameters improves the performance, even though the overall parameter efficiency might decrease. Within the 7B model, the Adapter method without freezing connector is remarkably efficient, utilizing only 3.060\% of trainable parameters and yet securing a high performance rate of 76.3\%, showcasing an optimal balance between parameter efficiency and model efficacy.

Additionally, using the IA3 method without freezing the connector can reduce the computing effort needed for training. With more trainable parameters, the model becomes more efficient, producing shorter and more accurate texts and thus requiring less computing power, even as the number of trainable parameters grows. In the case of the 13B model, the increase in the total number of parameters is not necessarily reflected in an increase in the percentage of trainable parameters to reach a good performance. According to Table ~\ref{tab:efficiency}, the IA3 method without a frozen connector yields significantly fewer trainable parameters compared to the Adapter and LoRA methods, but achieves the highest performance.

The details of the flops calculation are
: 

\smallskip
\inlineSubsection{Training Flops}  Since the computational cost of the backward pass is approximately twice as the forward pass, we modify the formula as:

\begin{equation}
    {\rm{\text{Train Flops}}} = (2{P_f} + 4{P_t})\times{N_t}
\end{equation}
where $P_f$ and $P_t$ represent the number of frozen and trainable parameters respectively, $N_t$ is the number of input and model-generated tokens. 

\smallskip
\inlineSubsection{Inference Flops}  We calculate the inference flops based on the following equation:

\begin{equation}
    {\rm{\text{Inference Flops}}} = 2\times({P_g} + {N_{layer}}d_{model}{N_t})
\end{equation}
where $P_g$ indicates \textit{non-embedding} parameters, $N_{layer}$ is the number of model's layers, and $d_{model}$ represents the dimension of the residual stream.

\label{appendix_D}

\section{Case Study}
\label{app:h}
\subsection{Hallucination Analysis}
\label{app:h_1}

\begin{figure*}[ht]
    \centering
    \includegraphics[width=0.95\textwidth]{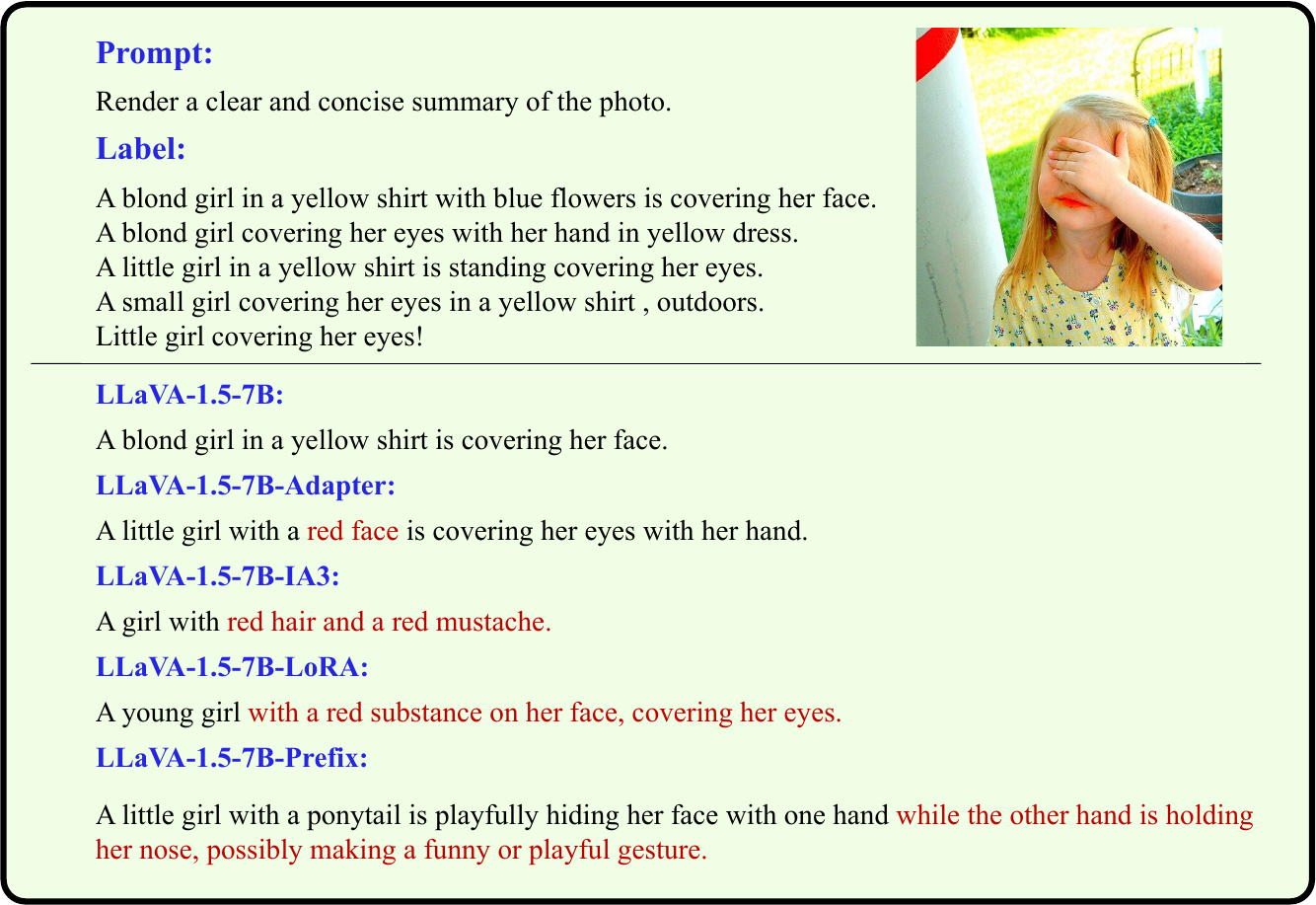}
    \caption{A qualitative hallucination example generated by LLaVA-1.5-7B and LLaVA-1.5-7B with various PEFT methods. LLaVA-1.5-7B generated hallucination-free text. After fine-tuning, all models generated hallucinations. }
    \label{fig:hal_case}
\end{figure*}

In this section, we explain how we tested the model's hallucination. As the first step, we employed LLaVA-1.5-7B to generate captions for all images in the Flickr30k test set in a zero-shot manner. Subsequently, we randomly sampled 1k captions and manually curated 100 correct captions without hallucinations. Then, we utilized the LLaVA-1.5-7B model fine-tuned with four PEFT methods on IconQA-txt to generate captions for these 100 samples. Thereafter, we manually annotate the fine-tuned model-generated outputs and count the number of hallucination samples.

One sample is selected, as illustrated in Figure~\ref{fig:hal_case}. In this example, the original LLaVA model delivers a hallucination-free description, accurately identifying the color and actions depicted in the image. On the other hand, the Adapter model incorrectly identifies the girl's face as red. The IA3 model inaccurately attributes a mustache to the girl and misidentifies her hair color as red. The LoRA model also fails to recognize the girl's hand and refers to a red substance, which is not present. As for the Prefix model, it attempts to provide a detailed description of the picture, including the mention of a ponytail and attributing emotions such as a "funny or playful gesture" to the subject. However, these emotions cannot be confirmed simply by viewing the picture. Additionally, the Prefix model presents more severe hallucinations compared to the others, as it "imagines" another hand that is not visible in the image. More examples at several epochs are illustrated in this section.

\begin{figure*}
    \centering
    \includegraphics[width=0.9\textwidth]{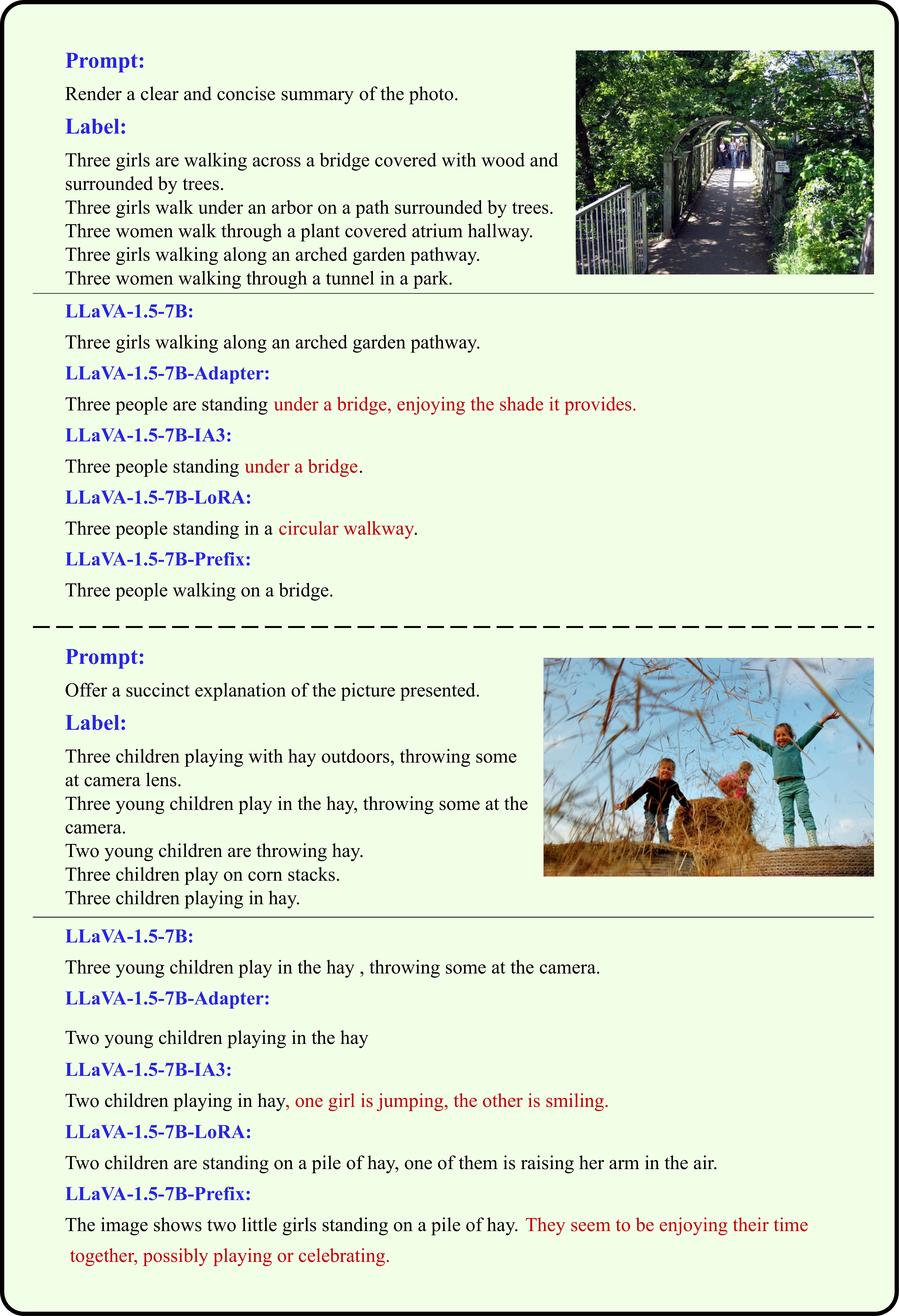}
    \caption{A randomly sampled hallucination results (from Flickr30K) of various PEFT methods' models trained on the IconQA-txt dataset for 3 epochs.}
    \label{fig:hlc_case_1}
\end{figure*}

\begin{figure*}
    \centering
    \includegraphics[width=0.9\textwidth]{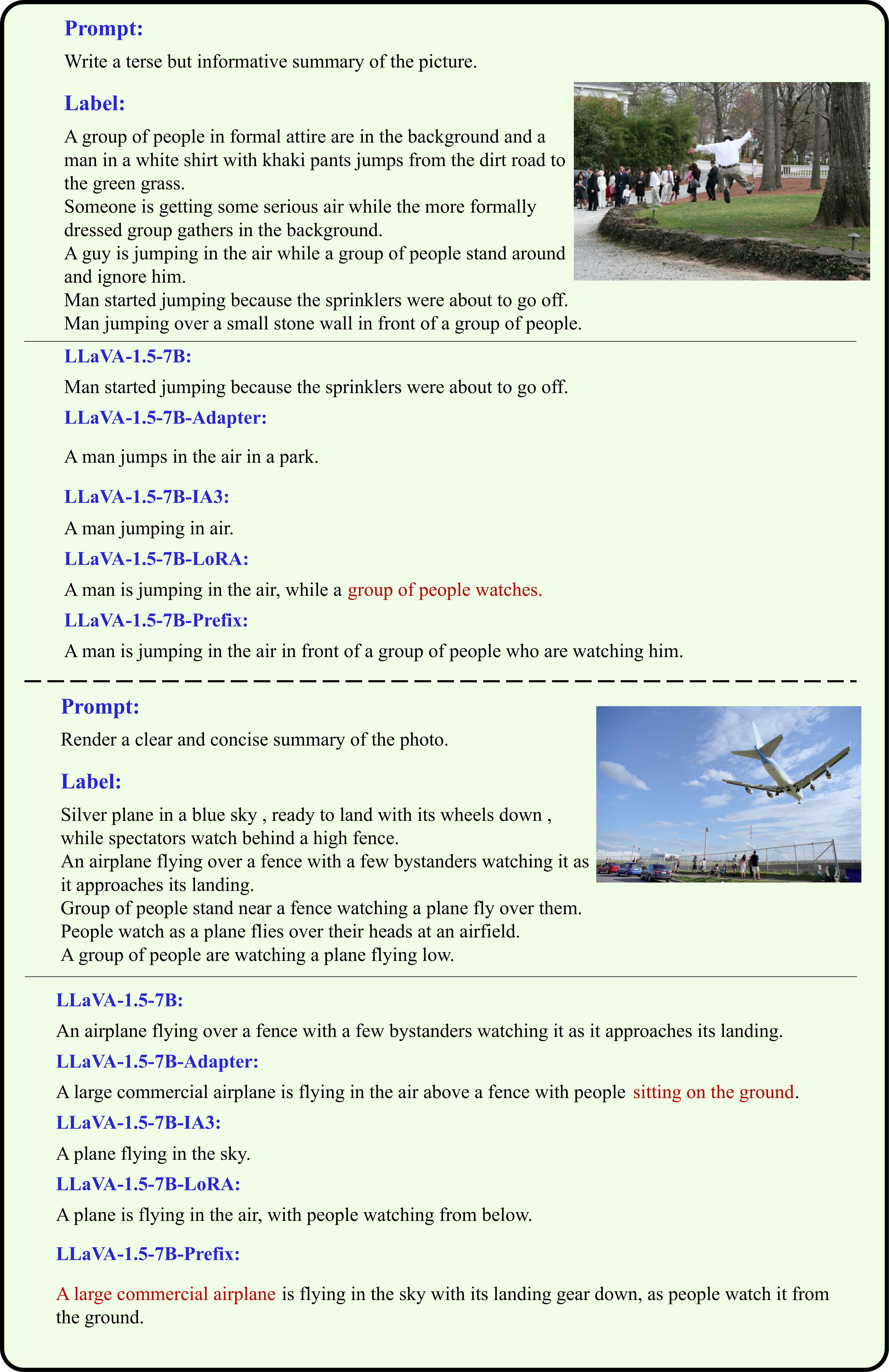}
    \caption{A randomly sampled hallucination results (from Flickr30K) of various PEFT methods' models trained on the IconQA-txt dataset for 6 epochs.}
    \label{fig:hlc_case_2}
\end{figure*}

\begin{figure*}
    \centering
    \includegraphics[width=0.9\textwidth]{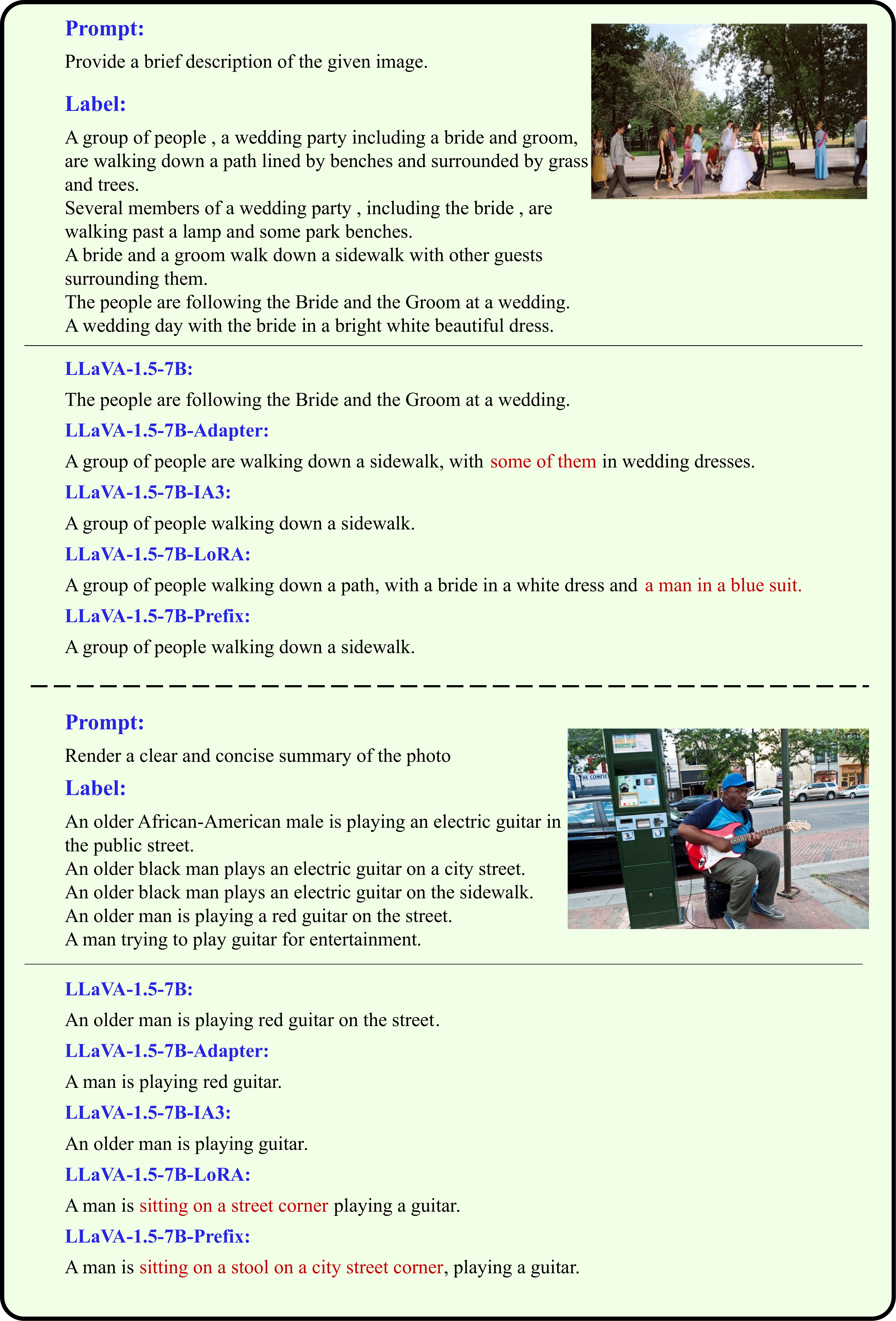}
    \caption{A randomly sampled hallucination results (from Flickr30K) of various PEFT methods' models trained on the IconQA-txt dataset for 9 epochs.}
    \label{fig:hlc_case_3}
\end{figure*}

\begin{figure*}
    \centering
    \includegraphics[width=0.9\textwidth]{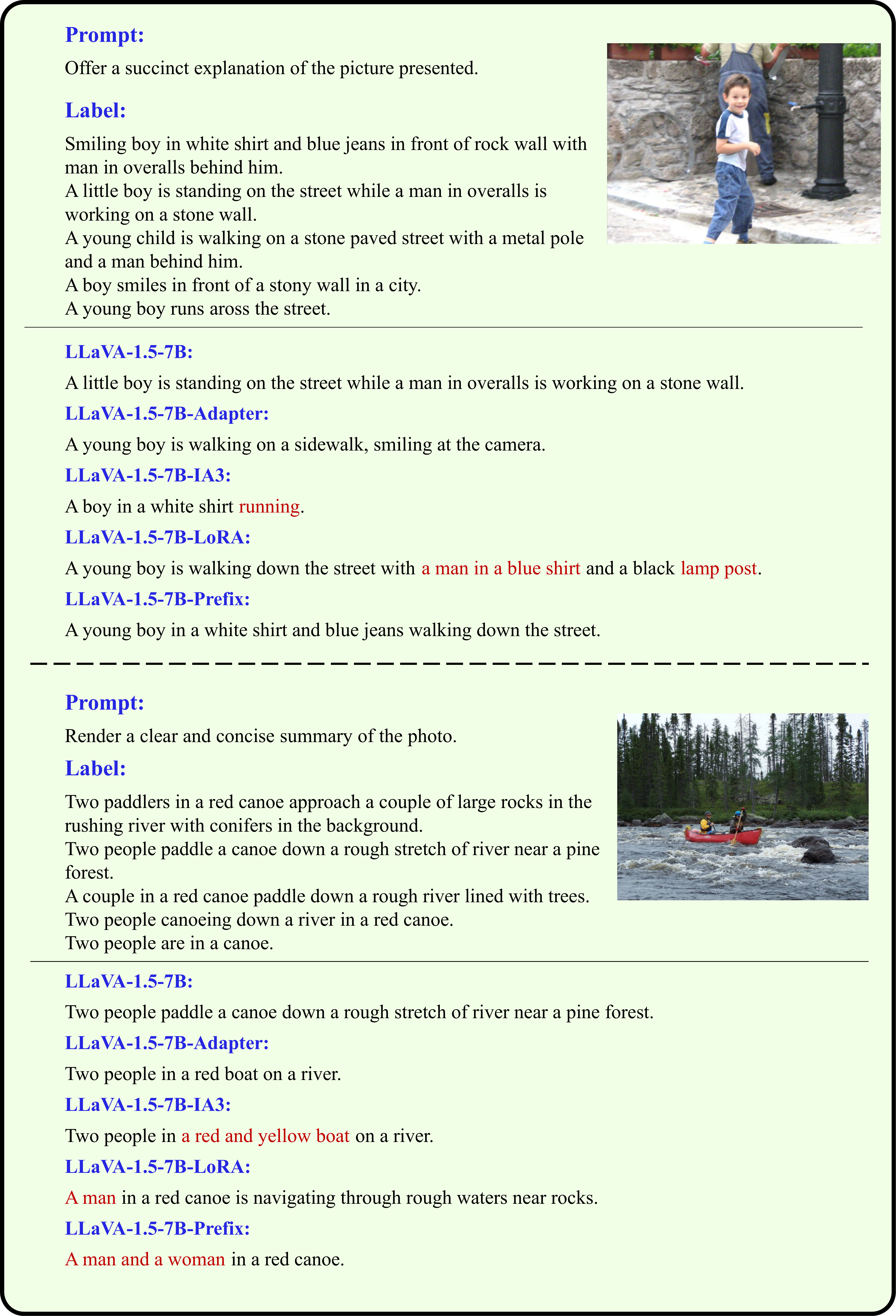}
    \caption{A randomly sampled hallucination results (from Flickr30K) of various PEFT methods' models trained on the IconQA-txt dataset for 12 epochs.}
    \label{fig:hlc_case_4}
\end{figure*}

\begin{table*}[h]
\footnotesize
    \centering
    \begin{tabular}{c|ccccc}
    \toprule
        Model & Method & Trainable parameters & Train flops & Inference flops & Performance\\
    \midrule
         \multirow{8}{*}{LLaVA-v1.5$_{7B}$}
         &Adapter       & 2.771\% & 1.526e+15 & 4.105e+10 & 74.6\\
         &-w/ connector & 3.060\% & 1.532e+15 & 4.109e+10 & 76.3\\
         &LoRA          & 4.332\% & 1.552e+15 & 4.092e+10 & 76.1\\
         &-w/ connector & 4.616\% & 1.561e+15 & 4.099e+10 & 76.7\\
         &IA3           & 0.009\% & 1.424e+15 & 4.028e+10 & 63.1\\
         &-w/ connector & 0.306\% & 1.248e+15 & 3.695e+10 & 76.3\\
         &Prefix        & 0.074\% & 1.496e+15 & 4.159e+10 & 68.1\\
         &-w/ connector & 0.371\% & 1.498e+15 & 4.154e+10 & 69.6\\
    \midrule
         \multirow{8}{*}{LLaVA-v1.5$_{13B}$}
         &Adapter       & 2.301\% & 2.898e+15 & 6.946e+10 & 77.2 \\
         &-w/ connector & 2.532\% & 2.908e+15 & 6.951e+10 & 77.7 \\
         &LoRA          & 3.615\% & 2.947e+15 & 6.942e+10 & 78.3  \\
         &-w/ connector & 3.842\% & 3.083e+15 & 7.127e+10 & 77.8  \\
         &IA3           & 0.007\% & 2.660e+15 & 6.716e+10 & 65.0  \\
         &-w/ connector & 0.243\% & 2.634e+15 & 6.667e+10 & \textbf{78.5}  \\
         &Prefix        & 0.061\% & 2.741e+15 & 6.838e+10 & 67.9  \\
         &-w/ connector & 0.297\% & 2.749e+15 & 6.840e+10 & 69.1  \\
    \bottomrule
    \end{tabular}%
    \caption{Efficient comparison across all types of settings. The number of trainable parameters, PEFT Flops, and Inference Flops are analyzed, considering whether to tune the connector with different PEFT methods on two models with different parameters. w/ connector: Tuning the connector.}
    \label{tab:efficiency}
\end{table*}

\subsection{Qualitative Illustrations}
\label{app:h_2}
\begin{figure*}
    \centering
    \includegraphics[width=0.7\textwidth]{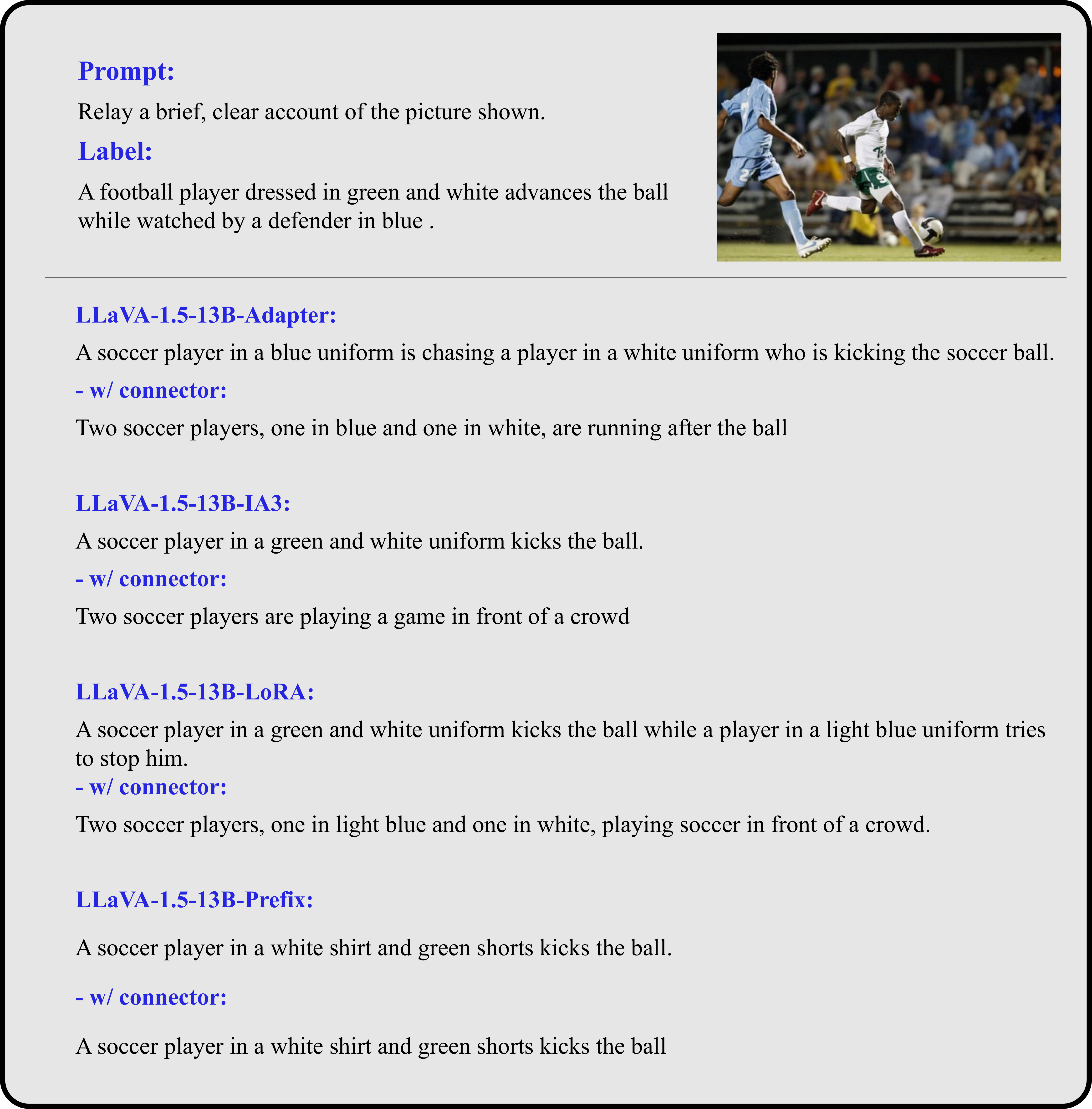}
    \caption{An example randomly chosen from the Flickr30K dataset. The outcomes produced by LLaVA-1.5-13B using different PEFT methods, each with the connector fine-tuned and frozen.}
    \label{fig:flickr_case_1}
\end{figure*}

\begin{figure*}
    \centering
    \includegraphics[width=0.7\textwidth]{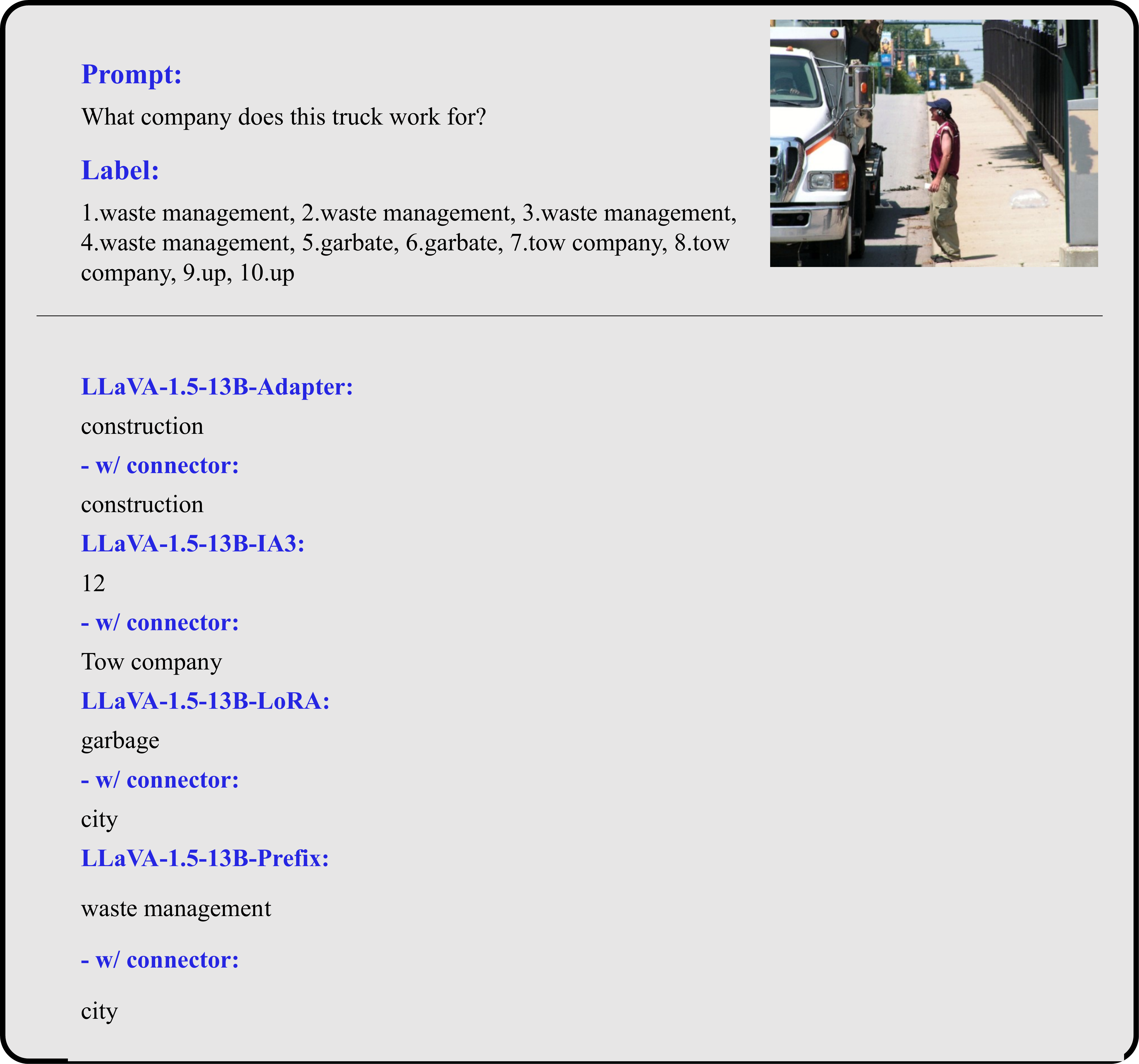}
    \caption{An example randomly chosen from the OKVQA dataset. The outcomes produced by LLaVA-1.5-13B using different PEFT methods, each with the connector fine-tuned and frozen.}
    \label{fig:okvqa_case}
\end{figure*}

\begin{figure*}
    \centering
    \includegraphics[width=0.7\textwidth]{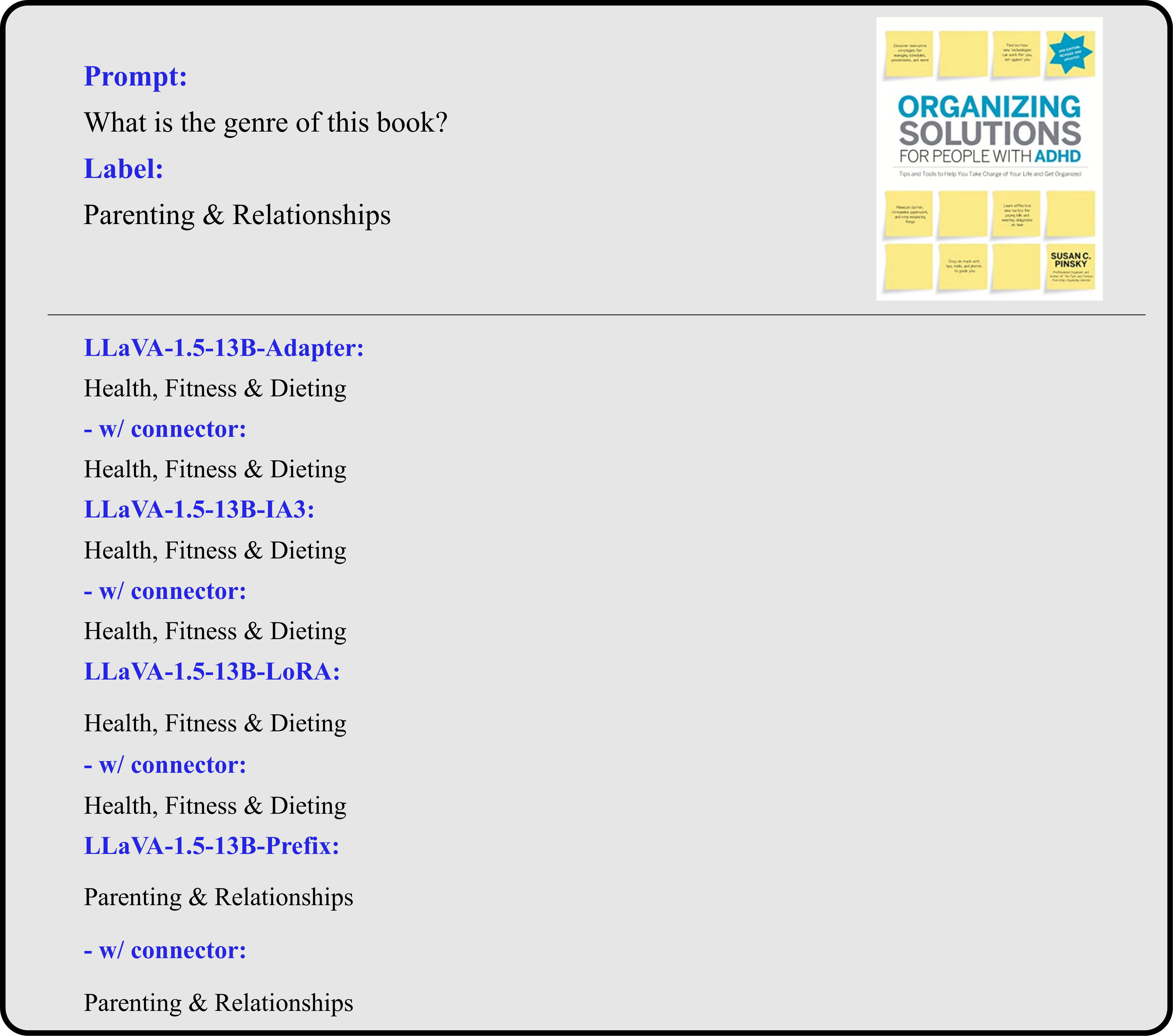}
    \caption{An example randomly chosen from the OCRVQA dataset. The outcomes produced by LLaVA-1.5-13B using different PEFT methods, each with the connector fine-tuned and frozen.}
    \label{fig:ocrvqa_case2}
\end{figure*}

\begin{figure*}
    \centering
    \includegraphics[width=0.7\textwidth]{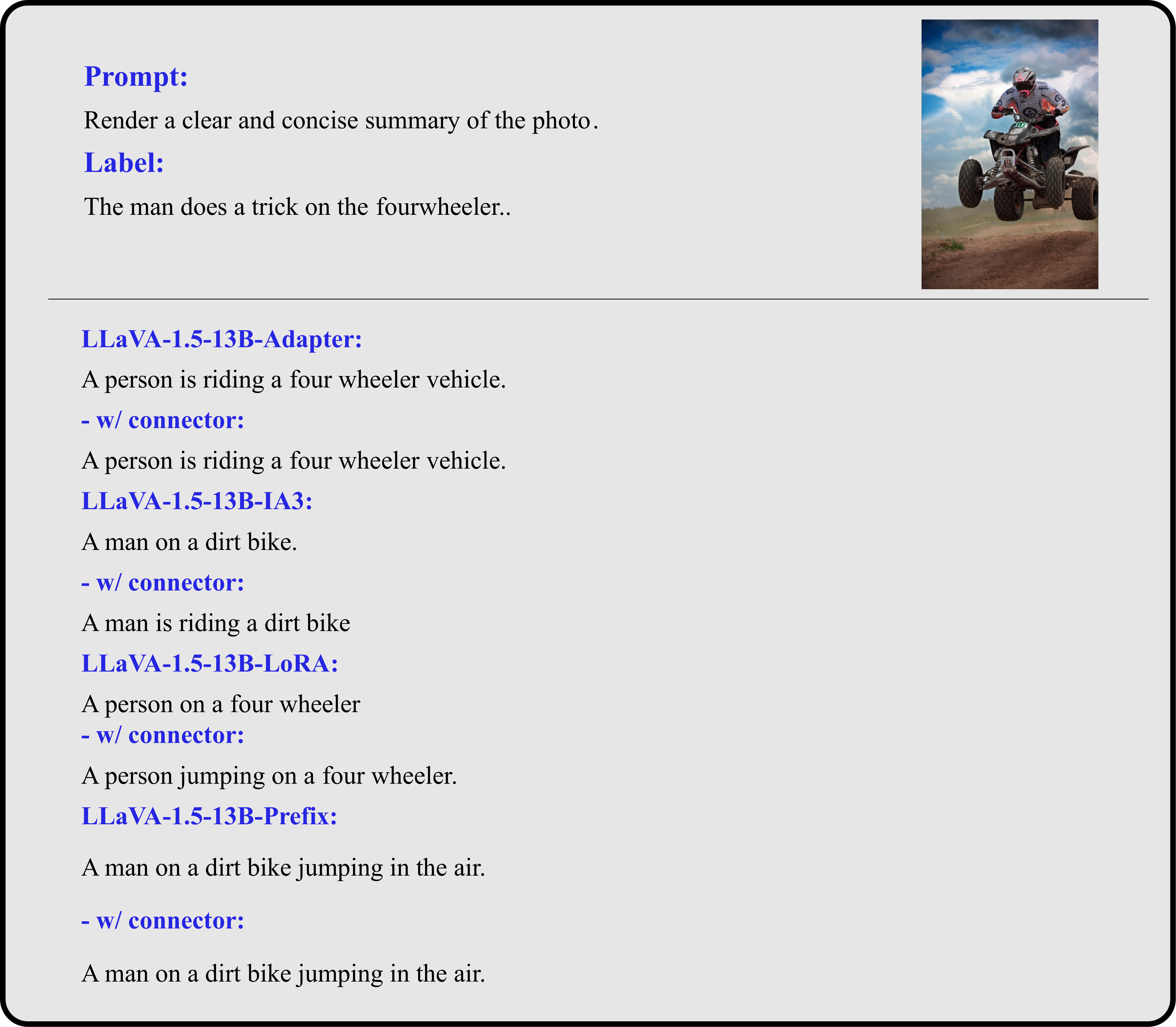}
    \caption{An example randomly chosen from the Flickr30K dataset. The outcomes produced by LLaVA-1.5-13B using different PEFT methods, each with the connector fine-tuned and frozen.}
    \label{fig:flickr_case2}
\end{figure*}

\begin{figure*}
    \centering
    \includegraphics[width=0.7\textwidth]{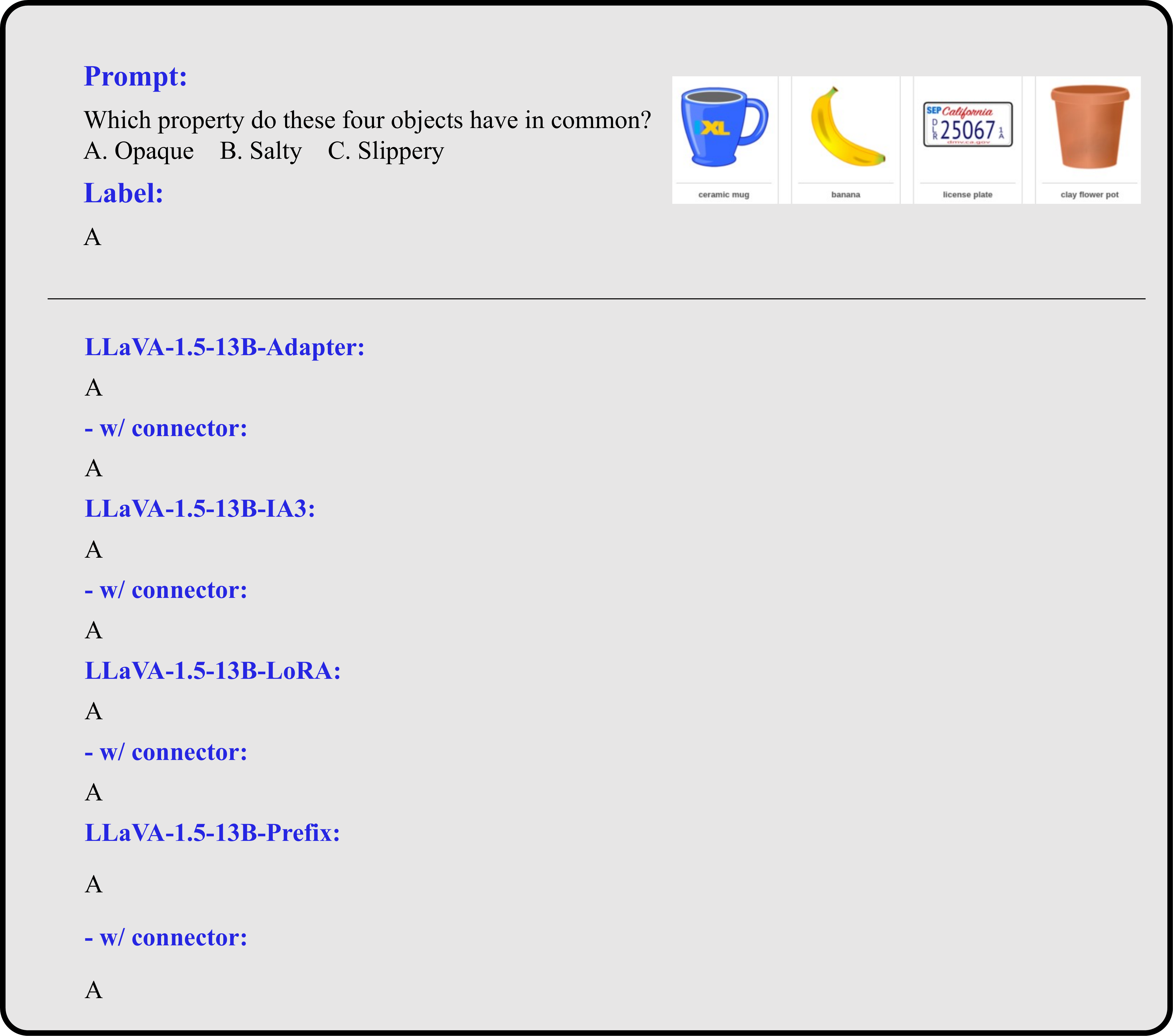}
    \caption{An example randomly chosen from the SQA (img) dataset. The outcomes produced by LLaVA-1.5-13B using different PEFT methods, each with the connector fine-tuned and frozen.}
    \label{fig:sqa_case}
\end{figure*}

\begin{figure*}
    \centering
    \includegraphics[width=0.7\textwidth]{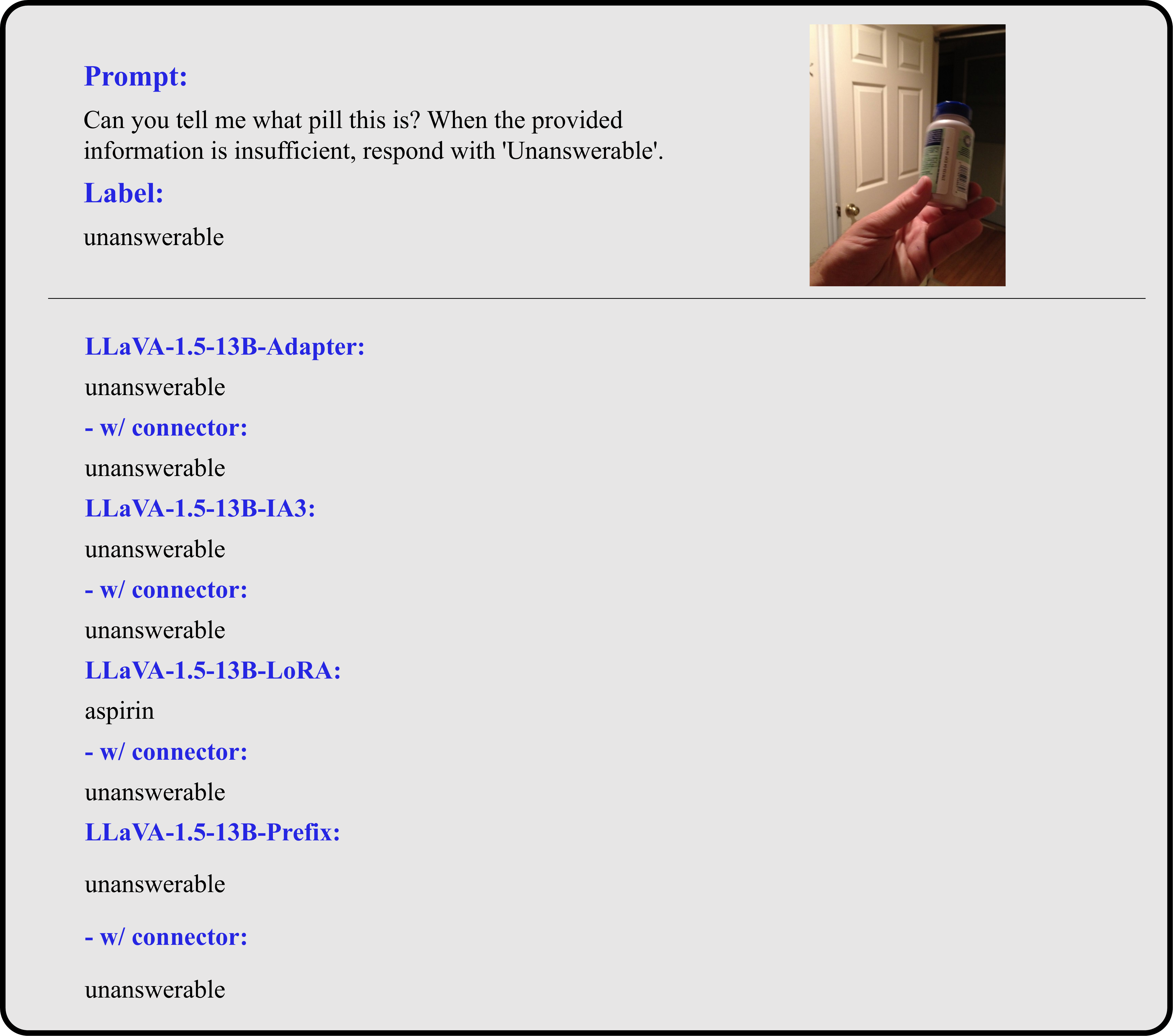}
    \caption{An example randomly chosen from the VizWiz dataset. The outcomes produced by LLaVA-1.5-13B using different PEFT methods, each with the connector fine-tuned and frozen.}
    \label{fig:vizwiz_case}
\end{figure*}

\begin{figure*}
    \centering
    \includegraphics[width=0.7\textwidth]{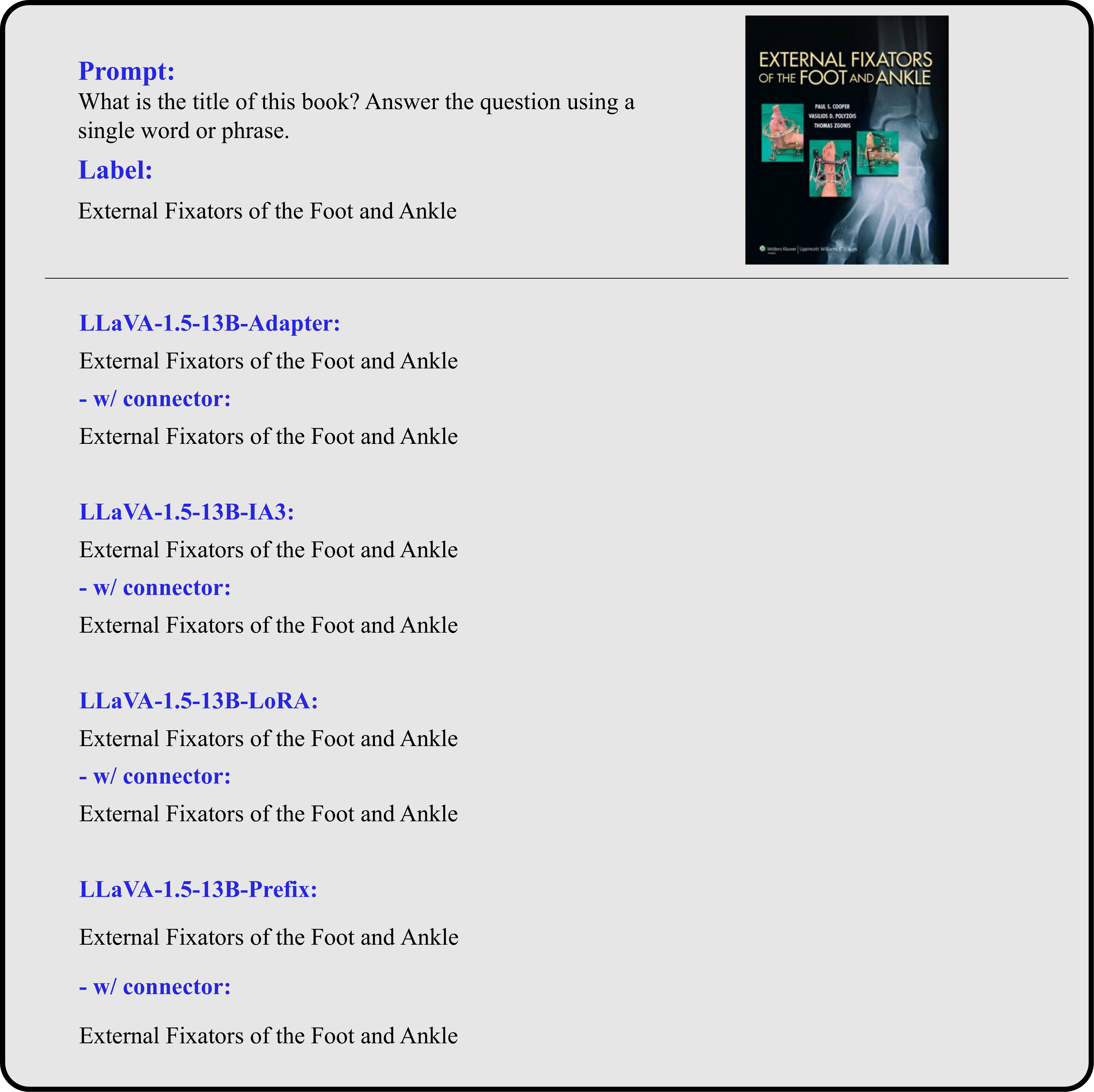}
    \caption{An example randomly chosen from the OCRVQA dataset. The outcomes produced by LLaVA-1.5-13B using different PEFT methods, each with the connector fine-tuned and frozen.}
    \label{fig:ocrvqa_case}
\end{figure*}

\begin{figure*}
    \centering
    \includegraphics[width=0.7\textwidth]{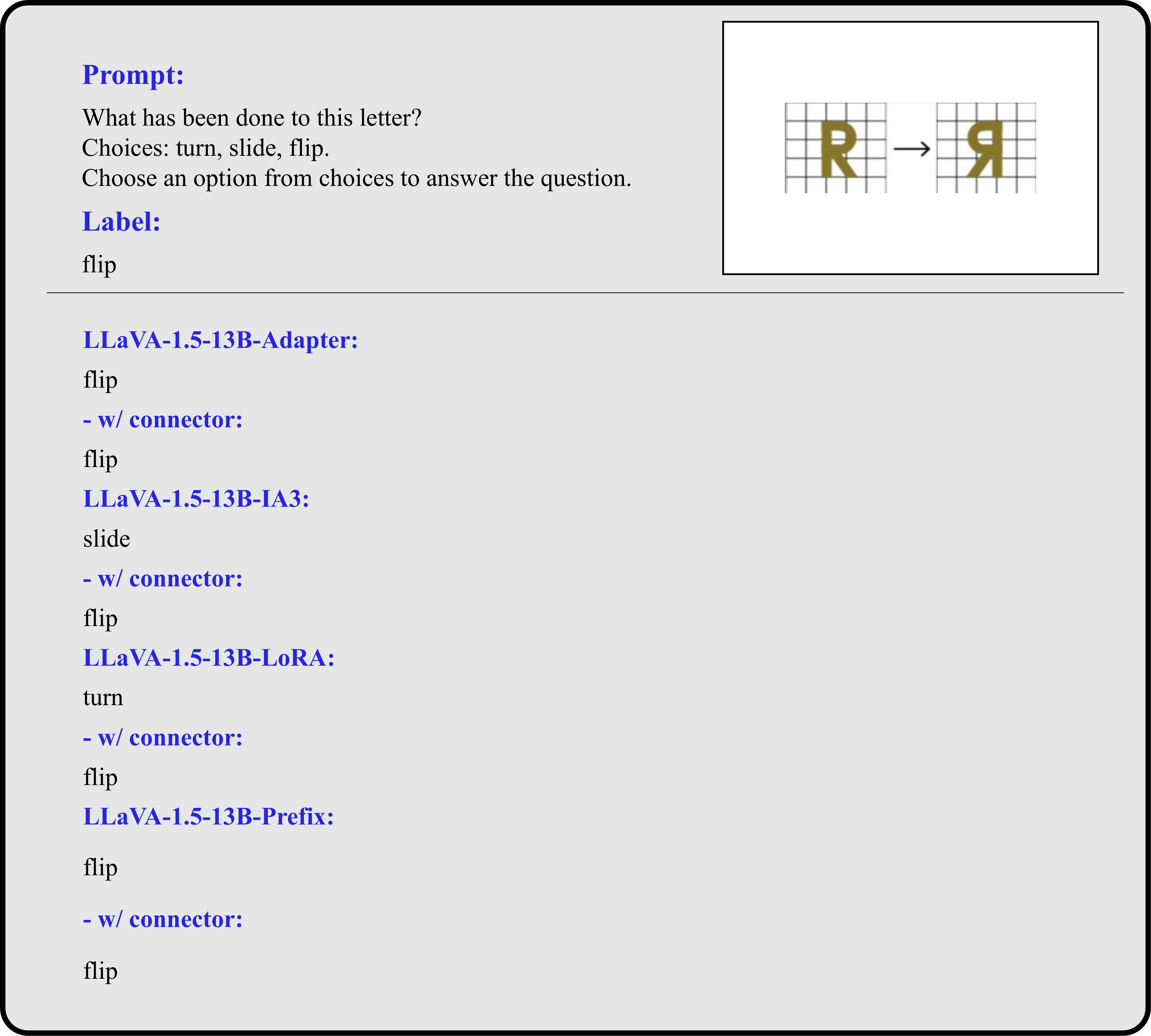}
    \caption{An example randomly chosen from the IconQA-txt dataset. The outcomes produced by LLaVA-1.5-13B using different PEFT methods, each with the connector fine-tuned and frozen.}
    \label{fig:txt_case}
\end{figure*}

\begin{figure*}
    \centering
    \includegraphics[width=0.7\textwidth]{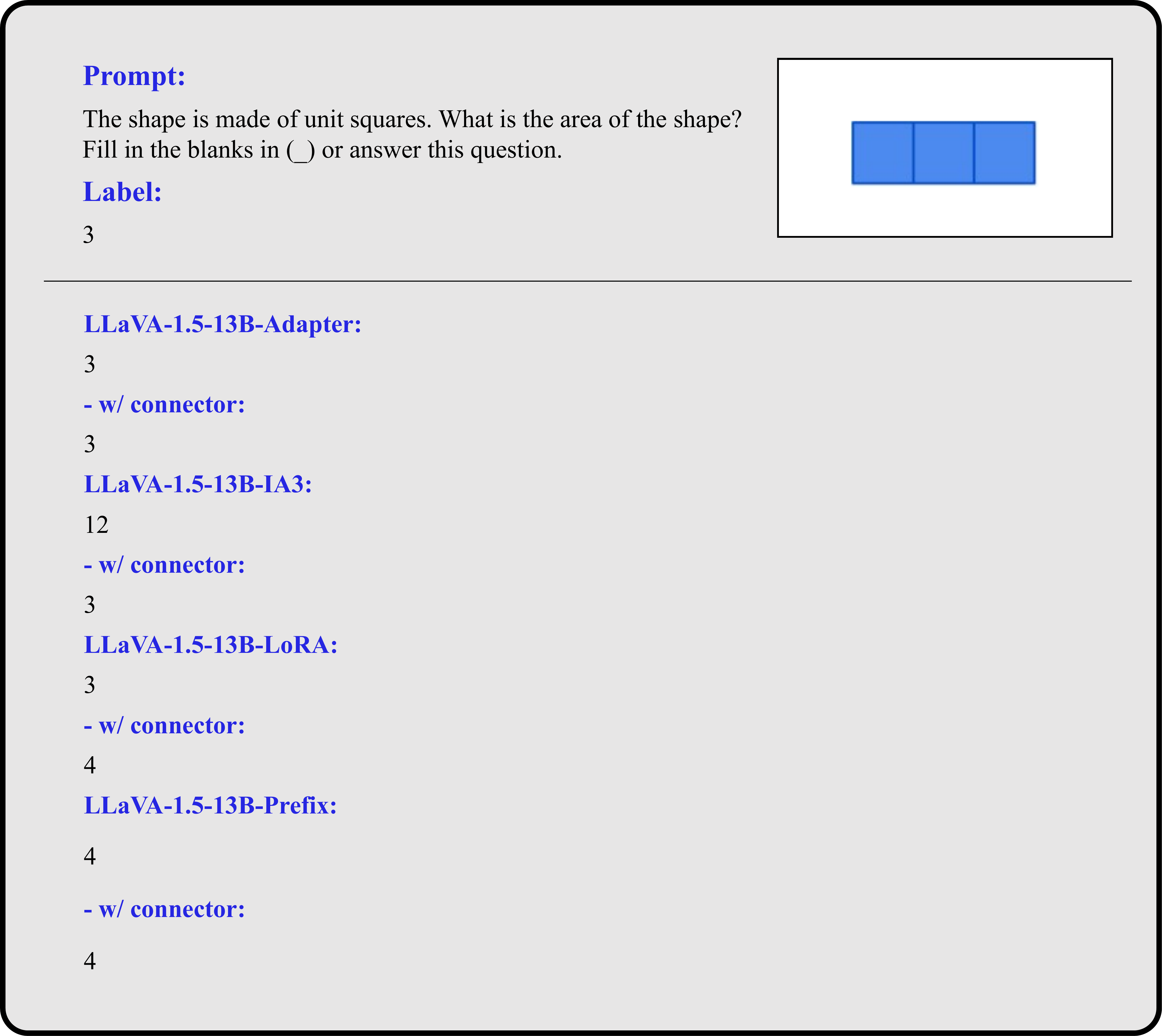}
    \caption{An example randomly chosen from the IconQA-blank dataset. The outcomes produced by LLaVA-1.5-13B using different PEFT methods, each with the connector fine-tuned and frozen.}
    \label{fig:blank_case}
\end{figure*}

\begin{figure*}
    \centering
    \includegraphics[width=0.7\textwidth]{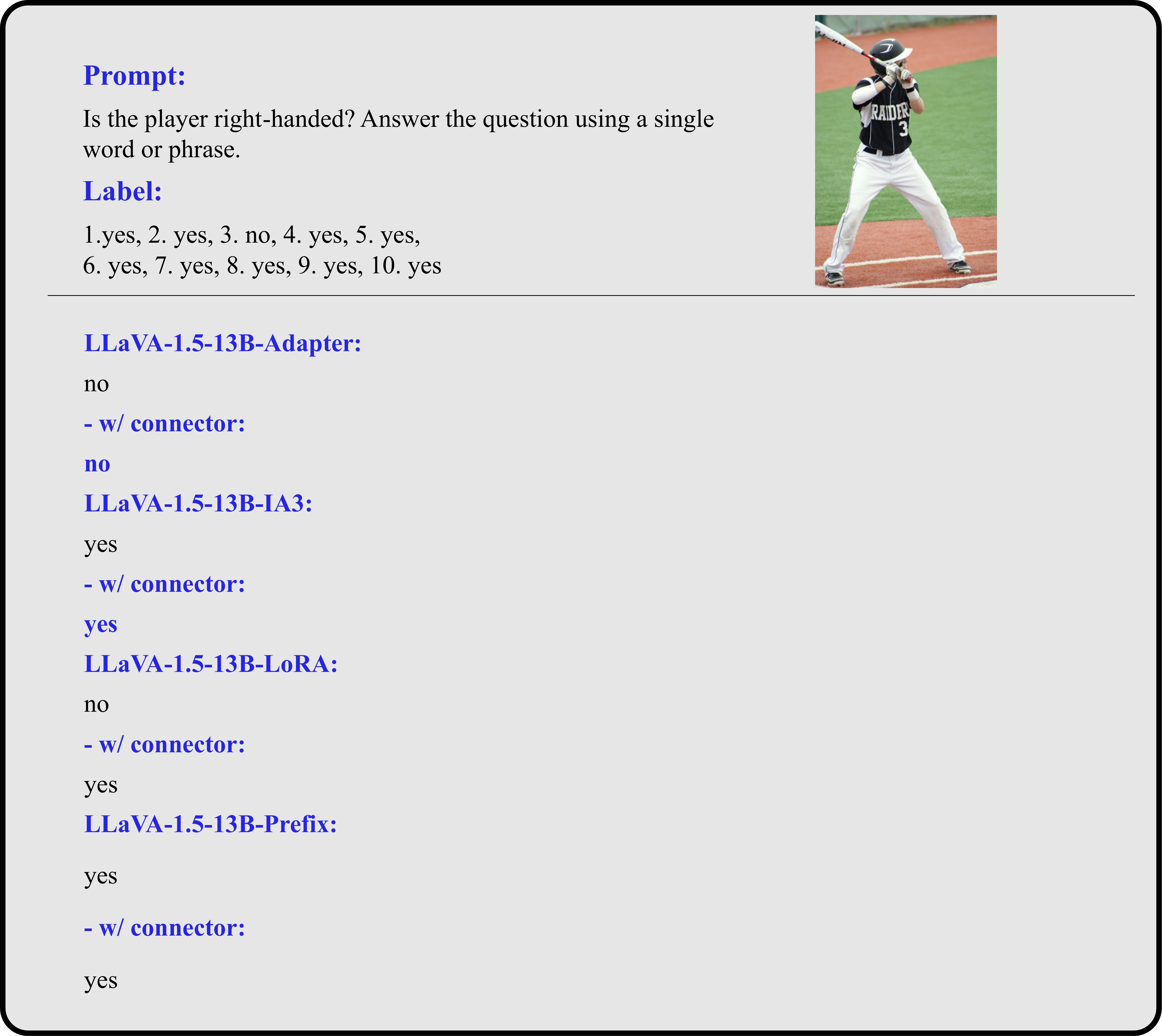}
    \caption{An example randomly chosen from the VQAv2 dataset. The outcomes produced by LLaVA-1.5-13B using different PEFT methods, each with the connector fine-tuned and frozen.}
    \label{fig:qua_case_1}
\end{figure*}

In this section, We randomly sampled several examples from each dataset and provided the original labels along with the outputs of various PEFT models. See Figures~\ref{fig:flickr_case_1} to~\ref{fig:qua_case_1}.

\end{document}